%% file: LocUNet_1004.tex
\documentclass[12pt, draftclsnofoot, onecolumn]{IEEEtran}

\input{header.tex}

\usepackage[section]{placeins}
\usepackage{url}
\usepackage{hyperref}
\usepackage{epstopdf}
\usepackage{enumerate}
\usepackage{algorithmicx}
\usepackage{algorithm}
\usepackage{amsmath}
\usepackage[noend]{algpseudocode}
\usepackage{float}
\usepackage{color}
\usepackage{makeidx}
\usepackage{bbm}
\usepackage{graphicx}
\usepackage{lipsum}
\usepackage{subcaption}
\usepackage[section]{placeins}
\usepackage{setspace}
\usepackage{lipsum}
\usepackage{adjustbox}
\usepackage{colortbl}
\usepackage{hhline}

\usepackage{multicol}
\definecolor{Gray}{gray}{0.88}
\definecolor{Grayy}{gray}{0.77}
\definecolor{Grayyy}{gray}{0.82}

\usepackage{multirow}

\def\BibTeX{{\rm B\kern-.05em{\sc i\kern-.025em b}\kern-.08em
		T\kern-.1667em\lower.7ex\hbox{E}\kern-.125emX}}

\newcounter{RonCounter}

\newcounter{CagkanCounter}

\newcounter{CagkanCounter2}

\newcounter{GCcounter}

\begin{document}
	\title{Real-time Outdoor Localization Using \\ Radio Maps: A Deep Learning Approach}
	
	%\author{
	%	\IEEEauthorblockN{%
	%		\c{C}a\u{g}kan~Yapar*, %
	%		Ron~Levie*, %
	%		 Gitta~Kutyniok, %
	%		and Giuseppe~Caire%
	%	}\\
	%	\IEEEauthorblockA{Technische Universit\"at Berlin,
	%		Germany\\ cagkan.yapar@tu-berlin.de, levie@math.tu-berlin.de, kutyniok@math.tu-berlin.de, caire@tu-berlin.de}\thanks{$^*$Equal contribution.}}%

	\author{\c{C}a\u{g}kan Yapar,~\IEEEmembership{Graduate Student~Member,~IEEE,}
	Ron Levie,\\ Gitta Kutyniok,~\IEEEmembership{Senior~Member,~IEEE,} and  Giuseppe Caire,~\IEEEmembership{Fellow,~IEEE} \\
% 		$^{\ddagger}$ Institute of Telecommunication Systems, 
% 		TU Berlin,\\
% 		$^{\j}$ Faculty of Mathematics, Technion - Israel Institute of Technology,\\
% 		$^{\dagger}$ Department of Mathematics,  
% 		LMU M\"unchen,\\
% 		%$^{\mathsection}$ Institute of Mathematics,  
% 		%TU Berlin,\\		
% 		$^{\S}$Department of Physics and Technology, University of Troms{\o}
% 	\date{}	
	\thanks{\c{C}a\u{g}kan Yapar is with the Institute of Telecommunication Systems, TU Berlin, 10623 Berlin, Germany (e-mail: cagkan.yapar@tu-berlin.de).}
	\thanks{Ron Levie is with the Faculty of Mathematics, Technion - Israel Institute of Technology, 3200003 Haifa, Israel (e-mail: levieron@technion.ac.il).}
	\thanks{Gitta Kutyniok is with the Department of Mathematics, LMU Munich, 80331 München, Germany, and also with the Department of Physics and Technology, University of Tromsø, 9019 Tromsø, Norway (e-mail: kutyniok@math.lmu.de).}
	\thanks{Giuseppe Caire is with the Institute of Telecommunication Systems, TU Berlin, 10623 Berlin, Germany (e-mail: caire@tu-berlin.de).}
	\thanks{A short version of this paper was presented in the IEEE International Conference on Acoustics, Speech, and Signal Processing (ICASSP 2022) in Singapore \cite{ICASSP}.
}
}
	\maketitle
	\vspace{-11.2mm}
	\begin{abstract}
	Global Navigation Satellite Systems typically perform poorly in urban environments, where the likelihood of line-of-sight conditions between devices and satellites is low. Therefore, alternative location methods are required to achieve good accuracy. We present LocUNet: A convolutional, end-to-end trained neural network (NN) for the localization task, which is able to estimate the position of a user from the 
 received signal strength (RSS) of  a small number of Base Stations (BS). 
 % In LocUNet, the user to be localized simply reports the measured RSS to a central processing  unit,  which  may  be  located  in  the  cloud. 
  %
  Using estimations of pathloss radio maps of the BSs and the RSS measurements of the users to be localized, LocUNet can localize  users with state-of-the-art accuracy and enjoys high robustness to inaccuracies in the estimations of radio maps. The proposed method does not require generating RSS fingerprints of each specific area where the localization task is performed and is suitable for real-time applications.
  %LocUNet obtains state-of-the-art results and enjoys high robustness. % to inaccuracies in the radio maps, which are used as input features ofthe proposed neural network. 
  Moreover, two novel datasets that allow for numerical evaluations of RSS and ToA methods in realistic urban environments are presented and made  publicly available for the research community. By using these datasets, we also provide a fair comparison of state-of-the-art RSS and ToA-based methods in the dense urban scenario and show numerically that LocUNet outperforms all the compared methods. 
   
% 	\ron{Are such long abstracts common at this journal? Is there some word limit?}\cagkan{200 was max., made it 200}
	\end{abstract}
	%\setstretch{1}
% 	\vspace{-9.2mm}
	
	%%%%%%%%%%%%%%%%%%%%%%%%%%%%%%%%%%%%%%%%%%%%%%%%%%%%%%%%%%%%%%%%%%%%%%%%%%%%%%%%%%%%%%%%%%%%%%%%%%
	\section{Introduction} \label{sec:Intro}

	The location information of a User Equipment (UE) is essential for many current and envisioned applications that range from emergency 911 services \cite{spect}, autonomous driving \cite{autonomousDriving}, intelligent transportation systems \cite{beyondGNSS}, proof of witness presence \cite{proofWitness}, 5G networks \cite{5G}, to social networks, asset tracking and advertising \cite{chal}, just to  name a few.

	In urban environments, Global Navigation Satellite Systems (GNSS) alone may fail to provide a reliable localization estimate due to the lack of line-of-sight conditions between the UE and the GNSS satellites \cite{GPSVehicle}. In addition, the continuous reception and detection of GNSS signals is one of the dominating factors in battery consumption for hand-held devices. It is thus necessary to resort to other complementary means to  achieve localization with the desired high accuracy. 

	In cellular networks, the position of UE can be estimated by using various statistics that the UE may report or the network can infer. The most prominent localization methods in the literature are based on Time of Arrival (ToA) \cite{TOA1,TOA2}, Time Difference of Arrival (TDoA) \cite{TDOA1}, Angle of Arrival (AoA) \cite{AOA1} and Received Signal Strength (RSS) \cite{RSS1} measurements.

%%%%%%%%%%%%%%%%%%%%%%%%%%%%%%%%%%%%%%%%%%%%%%%%%%%%%%%%%%%%%%%%%%%%%%%%%%%	
\vspace{-4.0mm}
	\subsection{Received Signal Strength (RSS)}\label{subseq:RSS}
	
     Received signal strength quantifies the received power, averaged over a certain time interval and frequency band, of the beacon slots broadcast periodically by base stations (BSs)  \cite{ZanellaSurvey,HeroRSS,bookRappaport}. For such measurements coherent reception is not necessary, and small scale fading fluctuations are alleviated due to the time-frequency averaging. Since the transmit power of the BS on its beacon signal slots is fixed and known, the received signal strength is a function of the pathloss of the propagation between the BS and the UE. In fact, received signal strength measurements of beacon signals are routinely performed by UEs and reported to the system as ``received signal strength indicators'' (RSSI). Reporting RSS information is a standard feature in  most current wireless protocols. For example, this is used to trigger handovers and to enable UE-BS association for load balancing purposes. Therefore, exploiting RSS information for localization is attractive since it is a feature already built-in in the wireless protocols and does not require any further specific hardware at the UE, whereas the time-based (ToA and TDoA) and angle-based (AoA) methods require high precision clocks and antenna arrays, respectively \cite{DBCorr,RSStoRange}.
%     \textcolor{red}{Furthermore, RSS is a robust quantity since the received beacon signal power is averaged  due to averaging, it is essentially immune from small scale frequency selective fading, to noise and interference levels.} \textcolor{black}{Confusing, interference levels of?} 
	 
	 \begin{remark}
	   For ease of exposition, throughout the paper we consider a cellular network scenario with BSs. Essentially, any wireless signal source with known location, e.g., WiFi-Hotspots, can be utilized.
	 \end{remark}

%%%%%%%%%%%%%%%%%%%%%%%%%%%%%%%%%%%%%%%%%%%%%%%%%%%%%%%%%%%%%%%%%%%%%%%%%%%%%%%%%%%%	
\vspace{-4.0mm}
	\subsection{Ranging-Based Methods}
	
	In ranging-based techniques, the distances between the UE and the BSs  are used to estimate the position of UE by lateration. Here, the distances are estimated  by using RSS or ToA measurements, based on a statistical signal attenuation or time delay model to estimate the distance between the UE and the BS. However, using such models is not appropriate in urban settings, since in practice the signal undergoes diverse propagation phenomena such as penetrations, reflections, diffractions, and wave guiding effects, due to the presence of obstacles in the environment. This leads to very large errors in the distance estimation (See e.g. \cite{MolischRT}). We call this phenomenon \emph{range  estimation  mismatch}. Several works, e.g. \cite{TOA1,TOA2,POCSgholami2011wireless,POCSHero,SDP,SDPR,ImpSDPR,BisecRob,Correntropy} proposed methods to mitigate the effects of obstructions in the range estimation mismatch.  Nevertheless, these methods do not directly use a complete model of the propagation phenomena, and only partially alleviate the aforementioned problem. 
	
    In particular, ranging techniques exploiting RSS are based on standard statistical models  where the pathloss is modeled as a deterministic function of the distance between the UE and the BS (radial function), with some (typically log-normal) random fluctuation. These techniques are particularly problematic in an urban setting \cite{MolischRT} where, for a specific propagation environment, the pathloss function is not at all a radial function (see for example Fig. \ref{fig:LocUNetResultsMild}). The usage of standard statistical models for RSS-based ranging has somehow generated the belief that RSS is an unreliable statistic. The matter of the fact is, that this is not a limitation of the RSS itself, but of the statistical models that relate pathloss with distance. It is therefore evident that in order to exploit RSS for positioning, a much more refined estimation of the pahtloss function for any specific propagation environment is needed.

%%%%%%%%%%%%%%%%%%%%%%%%%%%%%%%%%%%%%%%%%%%%%%%%%%%%%%%%%%%%%%%%%%%%%%%%%%%%%%%%%%%%%%%%%%%%%%%%%%%	
    \vspace{-4.2mm}
	\subsection{Fingerprint-Based Methods}\label{subseq:FPMethods}

	Fingerprint-based methods, as opposed to ranging-based methods, do not impose any modeling assumption on the signal strength propagation. Instead, they rely on offline generated extensive databases of the measured ``radio signatures'' of the signal at different locations. Given a reported signal fingerprint, these methods infer the location of the UE by ``looking up'' a location with a similar fingerprint from the database (radio map, cf. Section \ref{subseq:RMaps}). Many fingerprint types, such as visual, radio wave (e.g. RSS or the baseband transfer function with complex channel coefficients, i.e., the so-called Channel State Information (CSI)), or motion fingerprints, can be used for localization purpose \cite{FPSurvey}. In this paper we focus on RSS signatures which, as already argued before, are more easier to acquire (requiring time-frequency averaged non-coherent power measurements), more robust than complex baseband CSI to small-scale changes in the propagation environment, and much lower dimensional than CSI signatures. 
 Fingerprint-based methods are well-known for outperforming the ranging-based techniques in complex urban environments \cite{laitinen2011comparison}. Many machine learning (ML) techniques based on RSS fingerprints were applied for positioning in past work, e.g. \cite{DMMIMO_ML,MLLocSurvey}.

%%%%%%%%%%%%%%%%%%%%%%%%%%%%%%%%%%%%%%%%%%%%%%%%%%%%%%%%%%%%%%%%%%%%%%%%%%%%%%%%%
    \vspace{-4.2mm}
	\subsection{Physical Simulations of Radio Maps and RadioUNet}
	
	A major drawback of fingerprinting is the difficulty in generating and updating the radio signature database for the desired location. Fingerprinting requires a labor intensive, time consuming and expensive site surveying, which needs to be repeated as the environment changes. While this is possible for well-defined and somehow restricted indoor locations such as an airport hall, or a shopping mall, it is clearly difficult if not impossible to produce such data for an extended outdoor  area such as a whole city. Furthermore, the outdoor propagation medium changes rapidly due to moving vehicles. Therefore, radio signatures such as baseband CSI (sensitive to small scale effects such as multipath propagation and phase rotations) are extremely fragile. 
	
	When dealing only with RSS, a more feasible alternative for generating the radio map is using physical simulation methods such as ray tracing \cite{RayTracing,RayTracingNew,WinPropFEKO}, thereby bypassing the extensive measurement campaigns. Based on an approximate model of the physical signal propagation phenomena, such simulations yield very accurate predictions, especially when an accurate geometrical description of the urban environment is available, see e.g. \cite{DPM,IRTMANET}. Many previous works proposed using simulated radio maps for localization, e.g. \cite{DMMIMO_ML,locRtEnt,deviceInd,MaherMalaney,propModel,peopleEffect,FPWinPropDL,wolfle2002enhanced,MLWinPropMultPath}. 
	
	However, such simulations are computationally demanding and are thus not suitable for real time applications. Recently, an efficient deep learning-based method termed \emph{RadioUNet} \cite{RadioUNetTWC} (see Subsection \ref{RadioUNet}) was proposed by the authors of this paper. This algorithm enables the generation of high accuracy pathloss radio maps in much shorter time, and is thus adopted as an important building block of the current work.

%%%%%%%%%%%%%%%%%%%%%%%%%%%%%%%%%%%%%%%%%%%%%%%%%%%%%%%%%%%%%%%%%%%%%%%%%%%%%%%%%%%%%%%%%%%%%	
    \vspace{-2.2mm}
	\subsection{Machine Learning For Localization}
	
	Lately, several ML-based localization approaches have been proposed, e.g. \cite{DMMIMO_ML,locRtEnt,deviceInd,MaherMalaney,propModel,peopleEffect,FPWinPropDL,wolfle2002enhanced,MLWinPropMultPath,LTE_DL,deepCReg,GarauBurguera2020,RSSDNNFusion},  see the recent surveys \cite{MLLocSurvey,wirelessDLSurvey}. \textcolor{black}{To the best of our knowledge, none of the previous works considered applying convolutional neural networks on pathloss radio maps.} 
 % We note that the radio maps serve straightforwardly as a means to assess a likelihood for the location of the UE, for each BS, e.g., by simply comparing the reported RSS with the radio map estimate at the location at question. 
 We note that the radio maps may serve as a mean to assess the likelihood of a location for the UE with respect to a BS. \textcolor{black}{However, in order to compute such likelihood in a classical sense (i.e., not using a deep-learning training-based method), one would need to characterize the a posteriori probability of the location given the radio map and the RSS measurement. This requires in turns a statistical characterization that goes well beyond the simplistic pathloss models used in wireless communications.} 
	
	As opposed to the previous work, LocUNet does not require new signature generation (i.e., new RSS/CSI samples) for re-training when deployed in a new environment.  Moreover, LocUNet efficiently utilizes radio map estimations (See Section \ref{sec:OurMethod}) thanks to its convolutional design (See Remark \ref{remark:SpatialLocUNetKNN}).

    \vspace{-6.0mm}
	\subsection{State-of-the-Art Urban Localization in Current Commercial Use}
 \vspace{-2.2mm}

	The industry state-of-the-art on radio localization as implemented by systems such as Google Maps and Apple Maps are based on a multitude of sensor data, including GNSS, ``radio signatures’’ of all sort (e.g., WiFi SSIDs, cellular channel quality, detailed channel frequency response of the fading channels over the OFDM subcarriers, and so on), and these data are fused via some machine learning scheme. Since these practical implementations are ad-hoc industry intellectual property and not revealed in the open literature at a level of details sufficient to run accurate performance comparisons, it is virtually impossible to assess the performance of our proposed method against the ``real-world’’ state-of-the-art. 
	
	In contrast, it is important to notice here that our method is based uniquely on RSS measurements from known BSs/WiFi-Hotspots, which are collected by default by standard user devices since each device continuously measures the received signal strength from BS beacon signals. Furthermore, these measurements are very robust since BS beacon signals are easy to detect and do not require any calibration or accurate timing and frequency synchronization, as they are simply wideband power measurements. Our method based on such a simple setting, still achieves state-of-the-art results with respect to the previous works and could be  incorporated into some industry scheme by appropriate data fusion. This direction, however, goes beyond the scope of this paper and it is left for future work.

\vspace{-6.2mm}
\subsection{Our Contribution}	
\vspace{-2.2mm}
Our contribution can be summarized as follows. 1) We propose an accurate and computationally efficient localization method, merely based on RSS measurements, which does not necessitate additional signal processing or hardware (e.g., calibrated antenna arrays) at the user devices. 2) Relying on the recent developments in deep learning approximations  of ray tracing (RadioUNet \cite{RadioUNetTWC}) to generate pathloss radio map estimations, our method enjoys low run-time, which would allow real-time applications. 3) Using such radio map estimations and measured RSS values, the proposed LocUNet yields very accurate localization results, which are demonstrated by numerical experiments. The proposed method achieves about 5m of accuracy in mean absolute error, when trustworthy radio map estimates are available, which outperforms current state-of-the art. 4) The proposed method demonstrates very high robustness to the inaccuracies in the radio map estimates in a rigorously studied realistic setting. Moreover, the robustness of the method to out-of-distribution measurements, and to the inaccuracies in the location labels of the training set, are shown. 5) Extensive numerical comparisons with ToA ranging-based methods are presented, where LocUNet outperforms the state-of-the-art in different settings. 6) We justify our neural network design by a comprehensive ablation study. 7) Two synthetic datasets are introduced and set publicly available for the research community.

The rest of the paper is organized as follows. We present the preliminaries in Section \ref{sec:Preliminaries}. The proposed method is explained in Section \ref{sec:OurMethod}. Numerical results are presented in Section \ref{seq:Numerical}, and conclusions are drawn in Section \ref{seq:Conclusions}.

%%%%%%%%%%%%%%%%%%%%%%%%%%%%%%%%%%%%%%%%%%%%%%%%%%%%%%%%%%%%%%%%%%%%%%%%%%%%%%%%%%%%%%%%%%%%%%%%
\vspace{-4.2mm}
	\section{Preliminaries}\label{sec:Preliminaries}
        \vspace{-2.2mm}
	In this section, we present the preliminaries that serve as the building blocks of LocUNet. \vspace{-6.2mm}
	\subsection{Pathloss and Radio Map}\label{subseq:RMaps}
    	\vspace{-2.2mm}
	The \emph{pathloss} (or \emph{large-scale fading coefficient}), quantifies the loss of wireless signal strength between a transmitter (Tx) and receiver (Rx) due to large scale effects. The signal strength attenuation can be caused by many factors, such as free-space propagation loss, penetration, reflection and diffraction losses by obstacles like buildings and cars in the environment. In dB scale\footnote{$(\cdot)_{\rm dB} := 10\log_{10}(\cdot)$.}, pathloss amounts to $\textup{P}_{\textup{L}} = (\textup{P}_{\textup{Rx}})_{\rm dB}-(\textup{P}_{\textup{Tx}})_{\rm dB}$,
	where $\textup{P}_{\textup{Tx}}$ and $\textup{P}_{\textup{Rx}}$ denote the transmitted and received locally averaged power (RSS) at the Tx and Rx locations, respectively. Notice that ``locally averaged'' power is defined as the energy per unit time averaged over time intervals of the order of a typical transmission slot in the underlying protocol (e.g., the duration of a Resource Block in 5G NR standard \cite{zaidi20185g}) and over the whole system bandwidth. Hence, the effect of the small-scale frequency selective fading is averaged out and only the frequency-flat pathloss matters.   
A {\em radio map}  $R(\mathbf{x}_1,\mathbf{x}_2)$ defines the pathloss between any two points $\mathbf{x}_1$ and $\mathbf{x}_2$ in the plane. For fixed Tx position $\mathbf{x}_1 = \mathbf{x}_{\rm tx}$, the radio map is a function of the Rx position $\mathbf{x}_2 = \mathbf{x}_{\rm rx}$, i.e., it can be represented as a 2-dimensional image where the value of the pathloss $R(\mathbf{x}_{\rm tx},\mathbf{x}_{\rm rx})$ at each suitably discretized position $\mathbf{x}_{\rm rx}$ corresponds to a ``pixel'' of the image.

	\vspace{-4.2mm}
	\subsection{Radio Map Simulations}\label{subseq:RMapSims}
 \vspace{-2.2mm}
	
	One well-known class of numerical methods for solving Maxwell’s equations in far-field propagation conditions is {\em Ray Tracing} \cite{RayTracingNew}. In ray tracing, the signal is modeled as rays that are cast from the transmitter, travel in straight lines in homogeneous medium (like free space), and undergo rules of reflection, refraction and diffraction when the medium changes (e.g., when hitting an obstacle). 
	
	There are two main approaches for finding the ray paths in a given environment. Classical ray tracing searches for the paths between the transmitter and each receiver point, which necessitates very high computational time, with exponential dependency between the number of interactions and the complexity of the geometry (e.g., the number of buildings),  and a linear dependency between the computation time and the number of receiver pixels \cite{IRT}. The second approach, ray launching \cite{RayLaunching}, discretizes the propagation angle at the transmitter and launches rays in all directions. The rays interact with the environment as before, and terminate when a predetermined maximum number of interactions is reached or when the signal strength goes below a minimum value. The received signal strength is then computed as the sum of all rays at all receiver pixels (i.e., discretized locations). 

	In the following, we briefly explain the simulation models used in this paper, computed using the software WinProp \cite{WinPropFEKO}.

%%%%%%%%%%%%%%%%%%%%%%%%%%%%%%%%%%%%%%%%%%%%%%%%%%%	
	\subsubsection{Dominant Path Model (DPM)\cite{DPM}}\label{subsubsec:DPM}
	
	DPM is based on the assumption that the dominant propagation path from Tx to Rx must arrive via convex corners of the obstacles to Rx, and thus only diffractions are taken into account, omitting reflections and penetrations altogether.  
	%Namely, the model aims to find the propagation path from Tx to Rx only via diffractions. 
	The model assigns the pathloss of the ray with the largest received power to each receiver point. This is equivalent to finding the shortest free space path to each point, and computing the corresponding pathloss.
	The formula of the pathloss is given by some statistical model, based on the following parameters as inputs: The length of the path, a parameter that depends on the distance and visibility between the Tx and Rx, and a diffraction loss function, which depends on the angle of the diffraction and the number of previous diffractions in the path. 
	The model can also consider waveguiding pathloss effects when calculating the pathlosses of the paths, where reflection loss of the walls along the path as well as their distance to the path influence the value.
	This simplified model of the signal propagation allows a very efficient implementation with respect to more traditional ray tracing methods, while demonstrating good accuracy in many real-world urban propagation scenarios \cite{DPM}.
	
	%While providing accurate results, due to its simplicity, this model enjoys shorter run-times than IRT, which we describe in the following.
	
	\subsubsection{Intelligent Ray Tracing (IRT)\cite{IRT}}\label{subsubsec:IRT}
	
	This algorithm starts with a pre-processing step for the considered map, where the faces of the obstacles are discretized into regular tiles, their edges into segments, and the visibility relations among the centers of these elements are found.
	For the visible element pairs, the distance between their centers and the subtended angles are calculated. The computations of the paths, based on the ray launching approach, are then accelerated using the pre-computed visibility model.  
	The maximum number of interactions is set by the user.
	In our simulations, we set the maximum  number of interactions as two (at most one diffraction and one reflection). We set the tile length as 100 m, which results in having only one tile for most of the building walls and for all of the cars (cf. Section \ref{subseq:Dataset} for the detailed description of the dataset). High accuracy of this method was also proven by field measurements in several cities, see e.g. \cite{IRTMANET} and the references therein.

	\vspace{-6.2mm}
	\subsection{RadioUNet}
	\label{RadioUNet}
\vspace{-2.2mm}

	RadioUNet is a UNet-based \cite{UNet} pathloss radio map estimation method introduced in  \cite{RadioUNetConf,RadioUNetTWC}. In this paper we use the so called RadioUNet$_C$. The trained model takes as input the map of the city and the location of the BS (both encoded as B/W images) and returns an estimation of the corresponding radio map represented as a gray-scale image. RadioUNet demonstrates high accuracy, with root-mean-square-error of order of 1dB in various scenarios, and a run-time order of milliseconds on NVIDIA Quadro GP100 \cite{RadioUNetTWC}. RadioUNet is trained by supervised learning to match simulations of radio maps, using the RadioMapSeer Dataset \cite{DataPort}.

	%%%%%%%%%%%%%%%%%%%%%%%%%%%%%%%%%%%%%%%%%%%%%%%%%%%%%%%%%%%%%%%%%%%%%%%%%%%%%%%%%%%%%%%%%%%%
 \vspace{-6.2mm}
	\section{Pathloss Fingerprint-Based Localization with Deep Learning}\label{sec:OurMethod}
\vspace{-2.2mm}	
	In this section we present LocUNet -- a deep learning localization method based on estimations of pathloss radio maps. 
\vspace{-5.2mm}	
	\subsection{Proposed Method}\label{subseq:Method}
\vspace{-2.2mm}	
	Suppose that a UE with location coordinates $\mathbf{x}_U = (x_U, y_U)$ reports the strength of beacon signals (non-interfering identification signals), transmitted from a set of BSs $B_j$, $j=1,\ldots,J$, with known coordinates $\mathbf{x}_{B_j} = (x_{B_j},y_{B_j})$. We refer to such BSs as ``anchor BSs'' since they belong to the network of the wireless operator and, as such, their position is known.  Based on the relation between transmit/receive powers and the pathloss
	$(\textup{P}_{\textup{Rx}})_{\rm dB} = \textup{P}_{\textup{L}}+ (\textup{P}_{\textup{Tx}})_{\rm dB}$,
	the pathloss between the device and the BSs, $p_j$, $j=1,\ldots,J$, can be calculated, where we assume the small-scale fading effects are eliminated by averaging over time and system bandwidth. In our approach, the position of the UE is estimated based on the following information.
	\begin{enumerate}
		\item 
		The pathloss values $p_j$, $j=1,\ldots,J$, 
		\item
		The estimations of the radio maps for each anchor BS $R_j(x,y) := R(\mathbf{x}_{B_j}, (x,y))$, $j=1,\ldots,J$, computed via RadioUNet,\footnote{From now on it is convenient to represent the locations as pixel coordinates $(x,y) $ in the 2-dimensional plane.}
		\item
		The map of the urban environment, the locations of the anchor BSs (which are fixed and known). 
	\end{enumerate}
	The processing is typically performed in a centralized manne in the cloud, based on the RSS information reported by the UE. Alternatively, localized processing at the UE may also be considered, but in this case the UE must be aware of the BSs location (beyond their identity) and of the corresponding radio maps. At this point we do not distinguish between these two options, which are equivalent in terms of system performance (although may not be equivalent in terms of 
	protocol overhead, computation complexity at the UE, and privacy of the UE location with respect to the network control). 
	
	Our method, called LocUNet, computes an estimation of the UE location using a UNet variant. In order to input the above information (1)--(3) to the UNet,
	this is first represented as a set of 2D images.
	\begin{enumerate}
		\item
		The RSS values $\{p_j\}$ are converted to gray-level. Each pathloss $p_j$ is represented as a 2D image $P_j(x,y)$ of a constant value $p_j$, i.e., for each $j$ this is an all-gray uniform image, but the level of gray differs for different indices $j$. 
		\item
		Each radio map $R_j(x,y)$ is represented as a 2D image, with pixel values in gray-level. Radio maps are obtained by using RadioUNet, which takes the map of the urban environment, and the locations of BSs as input features \cite{RadioUNetTWC}.
		\item
		The map of the urban environment is represented as a binary black and white image, where the interior of the buildings are white (pixel value $=255$), and the exterior is black (pixel value $=0$). 
		\item
		The location of each BS $B_j$ is represented as a one-hot binary image, where the pixel at location $(x_{B_j},y_{B_j})$ is white, and the rest is black.
		
	\end{enumerate}
	Inputs (3) and (4) are optional for LocUNet as input features (Note that they are still needed to generate the input (2) via RadioUNet), and through numerical experiments we observe their negligible influence on the performance (See Section \ref{subsec:Impact_InpFeat}). 
	
	The first part of LocUNet is a UNet \cite{UNet}, with average pooling, upsampling + bilinear interpolation, and Leaky ReLU (with parameter 0.2) as the activation function. We call the output feature map of the UNet, $H(x,y)$ a \emph{quasi-heatmap}, as its value at a point $(x,y)$ in the map quantifies the likelihood of the UE to be located at this point, while it can take negative values, due to LeakyReLU being the activation function of the network. 
	
	The final layer of LocUNet computes the center of mass (CoM) $(\mu_x,\mu_y)$ of  $H(x,y)$,
	\begin{align*}\label{estimator}
    \mu_x = \frac{\sum_{x=1}^{256}\sum_{y=1}^{256}x H(x,y)}{\sum_{x=1}^{256}\sum_{y=1}^{256} H(x,y)} &, \quad
    \mu_y = \frac{\sum_{x=1}^{256}\sum_{y=1}^{256}y H(x,y)}{\sum_{x=1}^{256}\sum_{y=1}^{256} H(x,y)},
	\end{align*}
	where 256 is the number of pixels along each axis. 
 
 The architecture of LocUNet is summarized in Table \ref{table:LocUNet}. 

   \begin{table}[!t]
	\renewcommand{\arraystretch}{1}

	\centering
 \caption{\small Architecture of the first part of LocUNet (w/o final CoM layer). \emph{Resolution} is the number of pixels of the image in each feature channel along the $x,y$ axis. \emph{Filter size} is the number of pixels of each filter kernel along the $x,y$ axis.
		The input layer is concatenated in the last two layers before the CoM layer. $in = 10/11/15/16$, cf. Table \ref{table:activationInputsMSEMAE}.}
	\resizebox{\columnwidth}{!}{
		\begin{tabular}{c|c|c|c|c|c|c|c|c|c|c}
			\hline	
			\rowcolor{Grayyy} \multicolumn{11}{c}{\bfseries \quad \quad LocUNet}  \\
			\hline
			\rowcolor{Grayy} \bfseries Layer &\bfseries In &\bfseries 1 &\bfseries 2 &\bfseries 3 &\bfseries 4 &\bfseries 5 &\bfseries 6 &\bfseries 7 &\bfseries 8 &\bfseries 9\\
			\hline
			\cellcolor{Grayy} \bfseries Resolution & $256$ & $256$ & $128$ & $64$ & $64$ & $32$ & $32$ & $16$ & $16$ & $16$\\
			\hline
			\cellcolor{Grayy} \bfseries Channel & $in$ & $20$ & $50$ & $60$ & $70$ & $90$ & $100$ & $120$ & $120$ & $135$\\
			\hline
			\cellcolor{Grayy} \bfseries Filter size & $3$ & $5$ & $5$ & $5$ & $5$ & $5$ & $5$ & $3$ & $5$ & $5$\\
			\hline
			\rowcolor{Grayy} \bfseries Layer &\bfseries 10 &\bfseries 11 &\bfseries 12 &\bfseries 13 &\bfseries 14 &\bfseries 15 &\bfseries 16 &\bfseries 17 &\bfseries 18 &\bfseries 19\\
			\hline
			\cellcolor{Grayy} \bfseries Resolution & $8$ & $8$ & $4$ & $4$ & $2$ & $4$ & $4$ & $8$ & $8$ & $16$\\
			\hline
			\cellcolor{Grayy} \bfseries Channel & $150$ & $225$ & $300$ & $400$ & $500$ & $400+400$ & $300+300$ & $225+225$ & $150+150$ & $135+135$\\
			\hline
			\cellcolor{Grayy} \bfseries Filter size & $5$ & $5$ & $5$ & $5$ & $4$ & $5$ & $4$ & $5$ & $4$ & $5$\\
			\hline
			\rowcolor{Grayy} \bfseries Layer &\bfseries 20 &\bfseries 21 &\bfseries 22 &\bfseries 23 &\bfseries 24 &\bfseries 25 &\bfseries 26 &\bfseries 27 &\bfseries 28 &\bfseries 29\\
			\hline
			\cellcolor{Grayy} \bfseries Resolution & $16$ & $16$ & $32$ & $32$ & $64$ & $64$ & $128$ &   $256$ & $256$ & $256$\\
			\hline
			\cellcolor{Grayy} \bfseries Channel & $120+120$ & $120+120$ & $100+100$ & $90+90$ & $70+70$ & $60+60$ & $50+50$ & $20+20+in$ & $20+in$ & $1$\\
			\hline
			\cellcolor{Grayy} \bfseries Filter size & $3$ & $6$ & $5$ & $6$ & $5$ & $6$ & $6$ & $5$ & $5$ & -\\
			\hline

		\end{tabular}
	}
	 \label{table:LocUNet}
\end{table}

	We measure the accuracy of the proposed method over a given subset  
	$\mathcal{S}$ of the dataset by average Euclidean distance, i.e., mean absolute error (MAE), between the estimated UE location and the true UE location, given by 
	\begin{equation}\label{eq:MAE}
	L_\textup{MAE}(\mathbf{u},\tilde{\mathbf{u}}) = \frac{1}{|\mathcal{S}|} \sum_{k \in \mathcal{S}} ||\mathbf{u}^k - \tilde{\mathbf{u}^k}||_2,
	\end{equation}
	where $\tilde{\mathbf{u}^k}:=(\mu_x^k,\mu_y^k)$ and $\mathbf{u}^k:=(x_U^k, y_U^k)$ denote the LocUNet estimation and the corresponding true location of the $k$th instance of the dataset. 
	
	We also consider using the average squared Euclidean distance, i.e., mean squared error (MSE), as a loss function, which is the average squared 2D Euclidean distance
	\begin{equation}\label{eq:MSE}
	L_\textup{MSE}(\mathbf{u},\tilde{\mathbf{u}}) = \frac{1}{|\mathcal{S}|} \sum_{k \in \mathcal{S}} ||\mathbf{u}^k - \tilde{\mathbf{u}^k}||_2^2.
	\end{equation}
	 When using stochastic gradient descent for training, we take $\mathcal{S} = \mathcal{B}_m$ in one of the cost functions mentioned above, where $\mathcal{B}_m$ is the $m$-th mini-batch of the training set.

\vspace{-4.2mm}
	 
	\subsection{Datasets}\label{subseq:Dataset}
 \vspace{-2.2mm}
	We introduce two new datasets to allow for comparisons among RSS (pathloss) and ToA ranging-based algorithms.  These two new datasets, which we call \emph{RadioLocSeer Dataset} and \emph{RadioToASeer Dataset}, are available at \cite{DataPort} and \url{https://RadioMapSeer.GitHub.io/LocUNet.html}.

 \textcolor{black}{More details on the datasets can be found in \cite{DatasetPaper}.}

	\subsubsection{RadioLocSeer Dataset}\label{subseq:Dataset_RadioLocSeer}
	
	\begin{table}[!t]	
	\renewcommand{\arraystretch}{1}
 
	\centering
 \caption{RadioLocSeer Dataset parameters.}
	\scalebox{0.8}{
	\begin{tabular}{cc}
		%\hline
		\rowcolor{Gray}		\bfseries Parameter &  \bfseries Value\\
		%		\hline
	    Map size &  $256^2$ pixels\\
		%		\hline
		Pixel length &  1 meter\\
		%		\hline
		BS, UE height &  1.5 meter\\
		%		\hline
		Center carrier frequency & 5.9 GHz\\
		%		\hline
		Channel bandwidth &  10 MHz\\
		%		\hline
		Noise power spectral density &  -174 dBm/Hz\\
		%		\hline
		Transmit power &  23 dBm\\
		%		\hline
		Noise floor (NF) &  -134 dB\\

	\end{tabular}
	}
	 \label{table:RMapSeerPar}
  \vspace{-5.2mm}
\end{table}
	
	We introduce a dataset of city maps, BS locations, and the corresponding radio maps simulated by the ray tracing software WinProp \cite{WinPropFEKO} and by RadioUNet. The dataset is based on the test set of RadioMapSeer \cite{RadioUNetTWC}, which contains 99 maps and 80 BS per map. We report the parameters of the dataset in Table \ref{table:RMapSeerPar} (cf. \cite{RadioUNetTWC} Section III.B for details). 
	The maps are randomly split into 69 training, 15 validation and 15 test maps.
	
	For each map we randomly generate 200 UE positions and randomly pick 5 BS positions out of the 80  available ones. We repeat the latter step 50 times for each map, to represent different scenarios of BS deployment.
	The BS locations are chosen to be separated by at least 20 m.  Moreover, the BSs and UEs are restricted to lie in the middle $150 \times 150$ and $164 \times 164$ boxes about the center of the $256 \times 256$ grid of the RadioMapSeer images, respectively. We ensured that each UE can receive the signals of at least one BS with a power above the noise floor (NF) (cf. Table \ref{table:condRunTime} for the number of BSs in coverage in the test set.)

	For each of the above maps and BS location, we provide: 1) the radio map estimations via the RadioUNet model trained on DPM (see Subsection \ref{subseq:Scenarios}); 2) the simulated (by WinProp) radio maps obtained by DPM and IRT (cf. Sec. \ref{subseq:RMapSims}); 3) a second version of the ray-racing simulations including the effects of cars in the streets, where each map consists of 100 cars (if the roads of the corresponding maps are of sufficient length to fit them, otherwise the maximum number that can fit) of size $2 \times 5 \times 1.5$ (width $\times$ length $\times$ height), randomly generated near and along or perpendicular to roads; 4) the B/W images of the city maps, along with the coordinates of the considered BSs and the UEs.

	\subsubsection{RadioToASeer Dataset}\label{subseq:Dataset_RadioToASeer}

	We generated another dataset based on the dominant path model (cf. Subsection \ref{subsubsec:DPM}) via the software \emph{WinProp} \cite{WinPropFEKO}, of the same settings as in RadioLocSeer Dataset, which provides ToA information of the dominant paths, to allow for comparisons between RSS and ToA ranging-based methods in realistic urban scenarios.

	\begin{figure}[!t]
		\vspace{-2mm}
		\centering
		\includegraphics[width=0.35\linewidth]{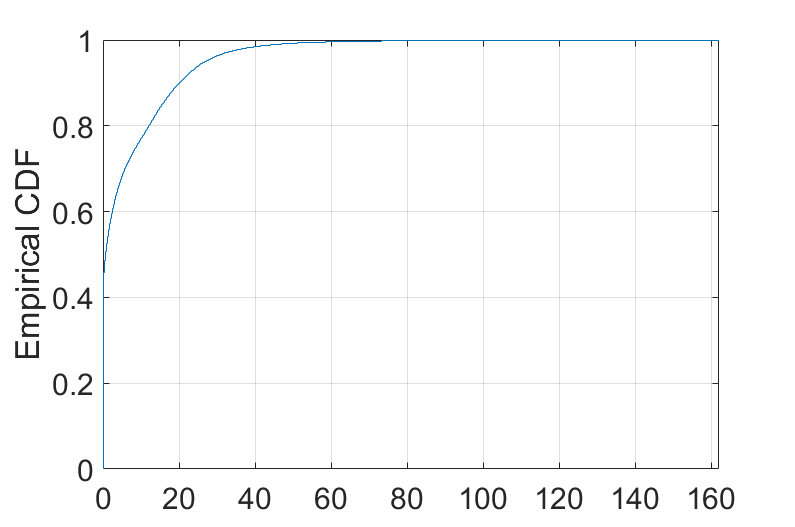}

		\vspace{-2mm}
		\caption{\small Empirical CDF of the difference between the direct and dominant path in meters}
		\label{fig:eCDF}
  \vspace{-9.2mm}
	\end{figure}

%	We argue that using this dataset serves as quasi-lower bounds for the errors of the ToA ranging-based methods, as explained next.

		We call the straight line connecting Tx and Rx, which may go through obstacles (in this paper, buildings and cars), the \emph{direct path}.
	The difference (error) between ranging with ToA measurement of the ray (range is calculated as: ToA $\times$ speed of light) and the direct path is called the \emph{NLOS bias}, which is by definition non-negative (Zero when the link is in LOS, positive otherwise). The empirical cumulative distribution function (CDF) of the NLOS bias of the dataset is shown in Fig. \ref{fig:eCDF}. We see that $47\%$ of the links in the RadioToASeer are in LOS.

	We argue that even though the dominant path may be a NLOS path, and hence introduces an NLOS bias, it is quasi-optimal to use the ToA of such paths in ToA ranging-based algorithms. First, if the dominant path is LOS, then it is by definition the shortest path and the bias is zero. Second, in the NLOS situation, the NLOS bias of a dominant path is lower than that of a potential free space path between the same Tx and Rx, that undergoes reflections to reach its destination (cf. Section \ref{subsubsec:DPM}). Thus, relying on ToAs of the dominant paths would yield better performances than the ToAs of the other free space paths.
	
	Notice that in NLOS conditions using the length (range) of the direct path (assuming building penetration) would also incur several problems. 
		First, we note that empirical evidence shows that loss due to penetration through a building is around $15-20$dB  \cite{585GHzRappaport}. In an urban scenario, the associated direct path of an NLOS link is usually subject to blockage by numerous buildings. For devices with realistic noise figures and SNR requirements, the received power of such a path may go easily below the detection threshold. 
		Furthermore, the bandwidth considered in the current paper does not allow for a well resolvability of the multi-path components. Indeed, in the considered setting, we have $\Delta \tau = 1/\textup{W} = 100$ns, so, when multiplied by the speed of light $c \approx 3 \times 10^8$ m/s, the difference between the paths should be at least $\Delta d = c \Delta \tau = 30$m for resolvability. In our dataset, we observe that around $92\%$ of NLOS direct paths don't satisfy this condition (cf. Fig \ref{fig:eCDF}), i.e., the difference of the length of the direct path and that of the dominant path is less than 30 m. 
		Finally, whenever the direct path penetrates an obstacle, it incurs an {\em excess delay}
		\begin{equation}
		t_\Delta = (\sqrt{\eps_r}-1)\frac{d_w}{c},
		\end{equation}
		where $\eps_r$ denotes relative electric permittivity and $d_w$ stands for the thickness of the wall. Assuming a total length of 3m of penetrated wall of concrete with permittivity 4, we get an excess delay of 10 ns, which further makes the ToA of the direct  and dominant path closer, deteriorating resolvability further: about $94\%$ of NLOS direct paths don't satisfy this. 
		
	   Hence, it is reasonable to ignore the direct paths and use the dominant paths instead. 
		
		Moreover, the dataset assumes exact ToA information, i.e., perfectly synchronized clocks at the BSs and UEs, and also no other additive noise on ToA measurements. Thus, all in all, evaluating the ToA ranging based-methods on this dataset yields essentially a best case of what is possible with ToA ranging in an urban environment. 

    \vspace{-2.2mm} 
   \begin{remark}To the best of our knowledge, there exists no publicly-available dataset of measured radio maps reporting consistent  signal strength and ToA with a fine resolution for the same environments. This is essential in order to make fair comparison of different methods, and we believe that our dataset will be useful for the research community in this area.
%    Notice also that the use of simulated datasets is widespread in the wireless communications community, as done in many previously published works addressing a variety of other problems (e.g., see
% \cite{MolischRT,AlkhateebBeamforming,AlkhateebBlockage,HeathInverse,DMMIMO_ML,locRtEnt,deviceInd,MaherMalaney,propModel,peopleEffect,FPWinPropDL,wolfle2002enhanced,MLWinPropMultPath,blockage, beamVehic, remoteChInference, mmBeamSel,adapDet, BICM, DLOFDM, CSIFeedback, ViterbiNet, Tabu, symRadioNet, energyAE}, just to name a few).  
   
   \vspace{-2.2mm}
   \end{remark}

    \vspace{-5.2mm}
   \subsection{Training}
   \vspace{-2.2mm}
	We perform supervised learning on the training set. We use \emph{Adam} optimizer \cite{Adam} with an initial learning rate of $10^{-5}$ and decrease the learning rate by a factor of $10$ after $30$ epochs. We set the total number of epochs for training as $50$ and batch size as $15$. To avoid overfitting, we pick the network parameters with the lowest validation error in the $50$ training epochs.
	 We used PyTorch \cite{pytorch} for the implementation\footnote{The code is available at \url{https://GitHub.com/CagkanYapar/LocUNet}. For reproducibility, see the compute capsule at \url{https://codeocean.com/capsule/7149386/tree}.}.

\vspace{-4.2mm}
\section{Numerical Results}\label{seq:Numerical}

	In this Section, we demonstrate the performance of LocUNet by numerical simulations. We assess the accuracy of LocUNet and the compared methods on the test set, namely, by MAE \eqref{eq:MAE} with $\mathcal{S} = \mathcal{T}$, where $\mathcal{T}$ is the test set (cf. Subsection \ref{subseq:Method}). 
	
	First, we discuss the sources of mismatch between a true and estimated radio maps and describe two scenarios in order to investigate the robustness of our method to such realistic inaccuracies. Afterward, we illustrate these scenarios and LocUNet's estimation results under them. Then, we compare the performance of the presented method with state-of-the-art fingerprint-based methods and study the effect of number of BSs in coverage. Subsequently, we study numerically the effect of out-of-distribution measurements, and the location errors in the position labels. Then, evaluations of state-of-the-art ToA ranging-based methods are presented, which proves the superiority of LocUNet in numerous settings. Next, we present the baseline/relevant previous methods, discuss their relation with LocUNet, and present their numerical evaluations. Finally, we present the ablation study of input features, activation function of the layer before the CoM layer, and the loss function for training, which justify our choices. 
	
	All of the compared algorithms with CPU implementation were run on an Intel Core i7-8750H, and LocUNet was run on a GPU of NVIDIA Quadro RTX 6000. We report the approximate run-times of the LocUNet and the compared algorithms (all implemented using MATLAB, run on CPU), as well. For the sake of completeness, we note that none of our implementations were meant to optimize the run-time, and thus the reported run-times reflect only an indicative assessment.
	%We note that none of our implementations were meant to optimize the run-time, and hence the reported times don't reflect a fair assessment.
\vspace{-4.2mm}	

	\subsection{Modeling of Mismatch in Radio Maps}\label{subseq:Scenarios}
	As we argued before (cf. Section \ref{subseq:RMaps}), the RSS values are obtained by integrating the signal strength measurements over time and the channel bandwidth, which effectively averages out the small scale frequency selective fluctuations. 
	
	Given this, we identify two sources of mismatch between the actual RSS measured at the UE and the estimated radio map input to LocUNet. These are: 
	
	\begin{enumerate}
	    \item The mismatch between the real propagation phenomena  and the simplified model which is used to estimate the radio maps. For example, a statistical model based on radial symmetry plus long-normal fluctuations exemplifies a very simplified model with large mismatch, whereas models that use the geometry of the environment (e.g. ray tracing methods or their accurate approximations like RadioUNet) will yield much better radio map estimations. 
	    \item The mismatch due to the inaccurate knowledge of the propagation environment. This inaccuracy can be decomposed into two classes: \textbf{1)} Imperfect knowledge on position, shape and material information of the objects, \textbf{2)} Changes in the environment due to the objects with mobility.
	    While the first class of inaccuracies could be alleviated by integrating public access city maps (e.g., \emph{OpenStreetMap} \cite{OpenStreetMap}) with refined information from cameras, LiDAR, and satellites, the second class of inaccuracy is not easily solvable since it would require a dynamic re-computation at a fast rate of the radio maps for the time-varying environment. 
	\end{enumerate}
	
	   Hence, in this work we focus on the inaccuracy due to mobile objects in the propagation environment, which we modeled as the presence of cars in the streets. In particular, these are unknown in the training phase of RadioUNet, and unknown at the run time of LocUNet. However, LocUNet can learn the discrepancy between the RadioUnet estimated radio map and the actual one during training, and (seamlessly) incorporate the statistics of this error in its location estimation. 

	In the following we designate two scenarios with increasing compliance with reality, encompassing the above mentioned sources of inaccuracies.

%\FloatBarrier

	\subsubsection{Nominal Scenario}\label{subseq:ScenariosDPM}
	In this  {\em optimistic} scenario, we assume that the real radio maps are \emph{exactly} governed by DPM  and the radio map estimates are obtained by RadioUNet, which was trained in a supervised fashion to estimate radio maps under DPM assumption, as well. Moreover, we assume that there are \emph{no addional unknown obstacles}, i.e., the \emph{exact} information about the propagation environment is available to RadioUNet. Hence, LocUNet enjoys having access to very high accuracy radio maps, where the inaccuracy of the available radio maps with respect to the true radio maps is solely due to the prediction error of RadioUNet, addressing the first mentioned source of inaccuracy.  

	\subsubsection{Robustness Scenario}\label{subseq:ScenariosIRT2carsDPM}
\begin{figure}[!t]
		\vspace{-2mm}
		\centering
		\includegraphics[width=0.45\linewidth]{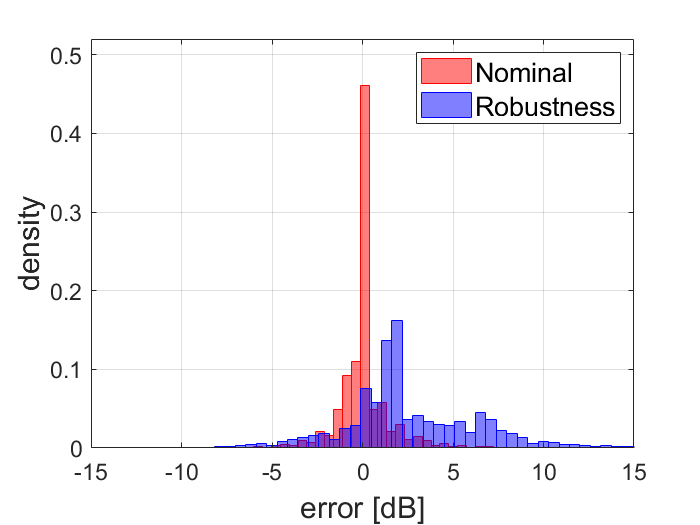}

		\vspace{-2mm}
		\caption{\small Histograms of the differences between the estimated (by RadioUNet trained on DPM simulations) and the assumed ground-truth pathloss measurements in RadioLocSeer Dataset. 
		% The Nominal and Robustness Scenarios use radio maps generated by DPM simulations, and IRT simulations with additional obstructions (cars) in the propagation environment, as surrogate ground-truths, respectively.}
            }
		\label{fig:hist}
  \vspace{-9.2mm}
	\end{figure}

	This scenario encompasses all the inaccuracies enumerated previously, which are likely to occur in the deployment of the method. That is to say, we take into account the potential mismatch between the real radio maps (above mentioned first source of error), which we simulate (with WinProp \cite{WinPropFEKO}) by IRT as the real-surrogate (from which the UE measures its pathloss), and the estimated radio maps, which are obtained by RadioUNet that was trained on DPM dataset. We also add the effects of unknown obstructions to the real-surrogate IRT simulations, addressing the second above mentioned source of error. We model the obstructions as cars (100 per map in general, cf. Sec. \ref{subseq:Dataset}). The root mean square error between the estimated and the simulated real-surrogate pathloss measurements are found to be 1.66 dB and 4.80 dB, for the Nominal and the Robustness Scenarios, respectively, whereas the mean absolute errors are 0.87 dB and 3.55 dB.  The histograms of the differences between the estimated and real-surrogate  pathlosses are shown in Fig. \ref{fig:hist} for each scenario.

    \vspace{-5.0mm}
    \subsection{Examples}\label{sec:Examples}
    \begin{figure}[!ht]
		\centering
		\begin{subfigure}{0.8\textwidth}
			\subfloat[][Tx 1 DPM]{\includegraphics[width=.18\textwidth]{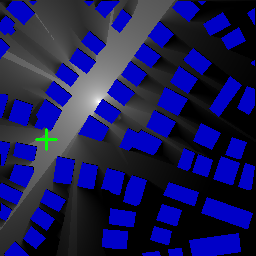}}\ %\quad
			\subfloat[][Tx 2  DPM ]{\includegraphics[width=.18\textwidth]{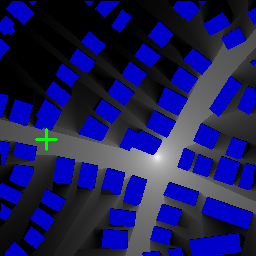}}\ %\quad
			%\subfloat[][IRT2]{\includegraphics[width=.24\textwidth]{Thr2FusedIRT2.png}}\quad
			\subfloat[][Tx 3 DPM]{\includegraphics[width=.18\textwidth]{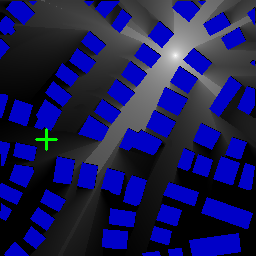}}\ %\quad%\\
			\subfloat[][Tx 4 DPM]{\includegraphics[width=.18\textwidth]{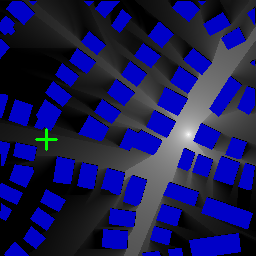}}\
			\subfloat[][Tx 5 DPM]{\includegraphics[width=.18\textwidth]{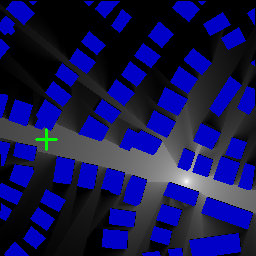}}\
			%\subfloat[][DPM with cars]{\includegraphics[width=.3\textwidth]{Thr2FusedDPMcars.png}}\quad
			%\subfloat[][IRT2 with cars]{\includegraphics[width=.3\textwidth]{Thr2FusedIRT2cars.png}}\quad
			%\subfloat[][IRT4 with cars]{\includegraphics[width=.3\textwidth]{Thr2FusedIRT4cars.png}}\quad
			%\caption*{Simulated models }
		\end{subfigure}
			\begin{subfigure}{0.8\textwidth}
			\subfloat[][Tx 1 IRT with cars ]{\includegraphics[width=.18\textwidth]{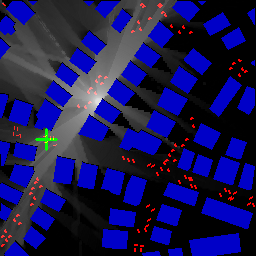}}\ %\quad
			\subfloat[][Tx 2  IRT with cars ]{\includegraphics[width=.18\textwidth]{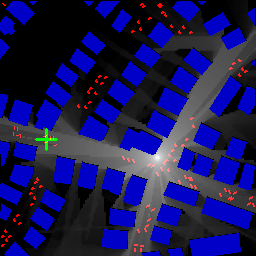}}\ %\quad
			%\subfloat[][IRT2]{\includegraphics[width=.24\textwidth]{Thr2FusedIRT2.png}}\quad
			\subfloat[][Tx 3 IRT with cars ]{\includegraphics[width=.18\textwidth]{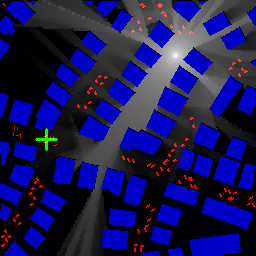}}\ %\quad%\\
			\subfloat[][Tx 4 IRT with cars ]{\includegraphics[width=.18\textwidth]{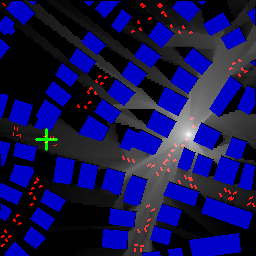}}\
			\subfloat[][Tx 5 IRT with cars ]{\includegraphics[width=.18\textwidth]{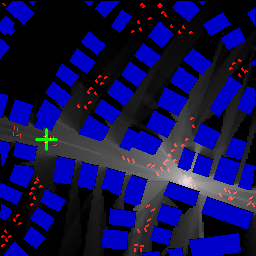}}\
		\end{subfigure}
		\begin{subfigure}{0.8\textwidth}
			\subfloat[][Tx 1 DPM via RadioUNet]{\includegraphics[width=.18\textwidth]{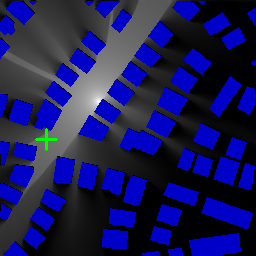}}\ %\quad
			\subfloat[][Tx 2 DPM via RadioUNet]{\includegraphics[width=.18\textwidth]{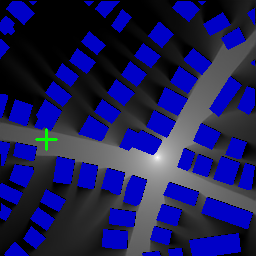}}\ %\quad
			%\subfloat[][IRT2]{\includegraphics[width=.24\textwidth]{Thr2FusedIRT2.png}}\quad
			\subfloat[][Tx 3 DPM via RadioUNet]{\includegraphics[width=.18\textwidth]{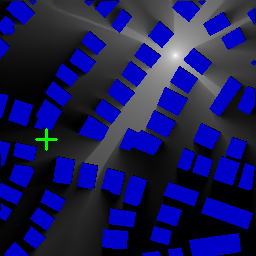}}\ %\quad%\\
			\subfloat[][Tx 4 DPM via RadioUNet]{\includegraphics[width=.18\textwidth]{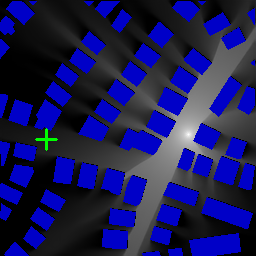}}\
			\subfloat[][Tx 5 DPM via RadioUNet]{\includegraphics[width=.18\textwidth]{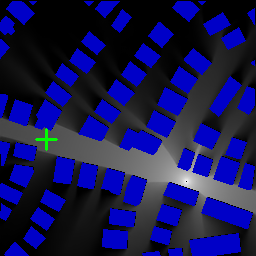}}\
			%\subfloat[][DPM with cars]{\includegraphics[width=.3\textwidth]{Thr2FusedDPMcars.png}}\quad
			%\subfloat[][IRT2 with cars]{\includegraphics[width=.3\textwidth]{Thr2FusedIRT2cars.png}}\quad
			%\subfloat[][IRT4 with cars]{\includegraphics[width=.3\textwidth]{Thr2FusedIRT4cars.png}}\quad
			%\caption*{Simulated models }
		\end{subfigure}
		\begin{subfigure}{0.7\textwidth}\centering
			%		\subfloat[][w/o city map DPMtoIRT2]{\includegraphics[width=.24\textwidth]{3_23_75buildsheatmapWOCityN.png}}\ %\quad
			\subfloat[][Nominal \newline positive heatmap]{\includegraphics[width=.24\textwidth]{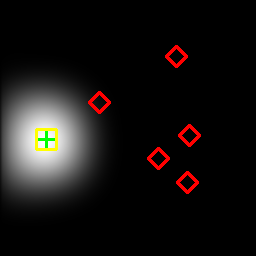}}\ %\quad
			\subfloat[][Nominal\newline negative heatmap]{\includegraphics[width=.24\textwidth]{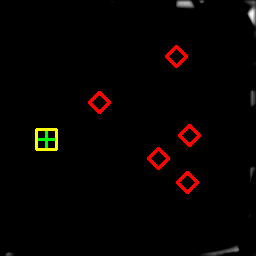}}\ %\quad
		\subfloat[][Robustness positive heatmap]{\includegraphics[width=.24\textwidth]{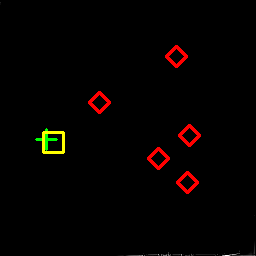}}\ %\quad
			\subfloat[][Robustness negative heatmap]{\includegraphics[width=.24\textwidth]{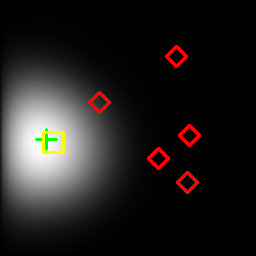}}\ %\quad
		\end{subfigure}
  \vspace{-2.2mm}
		\caption{\small A generic localization problem of our dataset, where the pathloss at the UE from each BS is moderate. \textbf{First row:} DPM true radio maps.  \textbf{Second row:} IRT with cars true radio maps. Cars are shown in red. \textbf{Third row:} DPM estimated maps by RadioUNet. \textbf{Fourth row:} LocUNet results for the two scenarios. The normalized positive (by max. pixel value) and the negative (by min. pixel value) heatmaps of LocUNet before CoM layer are shown in gray-level. Buildings are blue, Tx locations are marked with red diamonds. Estimated and true locations are marked with yellow square and green cross, respectively.} 
		\label{fig:LocUNetResultsMild}
  \vspace{-8.2mm}
	\end{figure}
\vspace{-2.2mm}
		In Fig.~\ref{fig:LocUNetResultsMild}, we show the results of LocUNet for a typical test map in the previously explained Nominal and Robustness Scenarios. 
In the first two rows, we show the radio maps for the considered 5 BS locations, which serve as the surrogate ground-truth (From top to bottom): DPM ground-truth (for the pathloss measurements at UE in the Nominal Scenario), IRT with cars ground-truth (for the Robustness Scenario). In the third row, the DPM estimations via RadioUNet are shown. %We marked the true position of the UE with a green cross.
In the last row, we show the heatmaps and the locations estimations (i.e., the CoM of the quasi-heatmap, marked with yellow squares) of the LocUNet for the considered two scenarios, where we partitioned the quasi-heatmap in two partial heatmaps, representing the positive and the negative values (normalized by its max. and min. pixel value, respectively), which we call the \emph{positive heatmap} and the \emph{negative heatmap}. The true location of the UE is marked with green cross.

We note that the CoM of the quasi-heatmap is invariant to the multiplication of the quasi-heatmap by $-1$.
		In each scenario, we observe that one of the heatmaps (the positive ones for the first scenario, and the negative one for the second scenario) serves as a \emph{belief map} about the location of the UE, while the other one as a \emph{disbelief map}. We see that the non-zero pixels (there are more of them in the Nominal Scenario) of the disbelief maps are concentrated about the pixels of the buildings and are distanced from the neighborhood of the estimated location.

	\vspace{-5.2mm}	
	\subsection{Comparison with Pathloss Fingerprint-Based Methods}
\vspace{-2.2mm}

	In Table \ref{table:MAERunTimesFP}, we present the MAE accuracy of LocUNet in the scenarios described in Subsection \ref{subseq:Scenarios}, and compare its performance with the competing fingerprint-based methods. The data is provided by the RadioLocSeer Dataset \ref{subseq:Dataset_RadioLocSeer}. LocUNet is trained using the whole dataset with 5 anchor BSs, where it is ensured that the UEs can decode the beacons of at least one BS (See Table \ref{table:condRunTime} for the number of BSs in coverage in the test set). Here, we use the radio map estimations of all 5 BS, even when the UE is not in the coverage of a BS (i.e., UE is located at a zero pixel of the true radio map of a BS). In order to study the effect of number of BSs in reception during the test phase, we also test our method on the test set where the number of BSs in coverage is not more than 3.

 \begin{table}[!t]
		\renewcommand{\arraystretch}{1}
		\centering
  \caption{\small Comparison with fingerprint-based methods. Average Euclidean distance (MAE) for 5 and 3 number of BSs, and approximate run-times of the evaluated methods.}
  \vspace{-2.2mm}
        \scalebox{0.8}{
        \begin{tabular}{cccc}
			\hline	
			\rowcolor{Grayy}  {\bfseries  Algorithm }& \bfseries 5Tx &\bfseries 3Tx & \bfseries ms\\
			\rowcolor{Grayyy} \multicolumn{4}{c}{\bfseries \quad \quad Nominal Scenario}  \\
			\hline
			kNN \cite{RADAR} (k=16, 40)  & $6.95$& $17.27$ & $\sim 20$\\
			\hline
			kNN \cite{RADAR} Cond. & $6.93$ & $16.56$ &$\sim 20$\\
			\hline
			Adaptive kNN \cite{adapKNN} (avg. k=$2.50,2.40$)  & $7.49$& $16.18$  &$\sim 20$\\
			\hline
			LocUNet &  $\mathbf{4.80}$&$\mathbf{10.70}$ &$\mathbf{\sim 5}$\\
			\hline
			\rowcolor{Grayyy} \multicolumn{4}{c}{\bfseries \quad \quad Robustness Scenario}  \\
			\hline
			kNN \cite{RADAR} (k=200,1500)  & $27.38$&$38.54$ &$\sim 20$\\
			\hline
			kNN \cite{RADAR}  Cond.  & $24.44$ & $32.83$ &$\sim 20$\\
			\hline
			Adaptive kNN \cite{adapKNN} (avg. k=$8.51,5.08$)  & $29.51$& $45.31$ &$\sim 20$\\
			\hline
			LocUNet  &  $\mathbf{13.14}$& $\mathbf{19.06}$&$\mathbf{\sim 5}$\\
			\hline  
		\end{tabular}
		}
		 \label{table:MAERunTimesFP}
   \vspace{-7.2mm}
\end{table}

	We compare with two fingerprint-based methods, namely, \emph{k-nearest neighbors (kNN)} method \cite{RADAR} and an adaptive kNN variant \cite{adapKNN}. We determined the $k$ parameters of the kNN algorithm by coarse grid-search for both of the 3 and 5 BS settings, as the one which yields the best performance in the training set. We also considered determining individual $k$ parameters for each number of BSs in coverage (See the following Subsection \ref{subseq:CondPerf} for more) in the same way, and report the overall performance of this approach under the name \emph{kNN Cond.} In the kNN methods, we have not excluded the pixels of buildings from consideration when finding the $k$ neighbors (when a BS is not in coverage), and doing otherwise did not yield improvements.

We observe that LocUNet provides the best accuracy among the fingerprint based methods in all the scenarios, and LocUNet is especially good at dealing with the inaccuracies of the radio map estimations in the realistic setting, as witnessed in the results of the Robustness Scenario (Sec. \ref{subseq:Scenarios}). Thus, the results in this scenario are a more meaningful indicator for the performances of the compared algorithms. LocUNet outperforms the best performing competitor algorithm (kNN) by about  11m and 14m in MAE in the cases of 5 or 3 BSs per map.

\begin{remark}\label{remark:SpatialLocUNetKNN}
We attribute the success of LocUNet in accuracy and its robustness to the inaccuracies in radio map estimations to its convolutional nature, which takes fully into account the neighborhood/spatial relations in the radio map estimations, whereas the kNN method makes use of the spatial information only when the CoM of the k-nearest-neighbors are found in the final step. The kNN method can hence be seen as a convolutional method with convolution kernels of 1x1 pixels. Note that both kNN and our method perform CoM as the last step and the found k-nearest-neighbors can be interpreted as a heatmap. However, the k-nearest-neighbors are solely determined by assigning each pixel a distance in the signal space of the RSS values under a metric (we took the usual ``Euclidean'', the sum of the squared residuals, and our experiments with the ``Manhattan'' metric did not yield improvements), which disregards any spatial relations of the pixels. LocUNet also compares the RSS measurements (represented via the constant image channel $p_j$) to the RSS values at each of the pixels (via the input radio map). However, LocUNet also utilizes the spatial information by propagating the RSS information from the spatial neighbors of each point via the convolutions.
\end{remark}

	\subsubsection{Conditional Performance of the Compared Methods}\label{subseq:CondPerf}

Previously, we have evaluated LocUNet and the kNN algorithms for 5 BS deployments or for subsets of 3 BS deployments of them. However, not all BSs are necessarily in coverage (We only ensured each UE is at least in the coverage of one BS, cf. Sec. \ref{subseq:Dataset}), as happens often in the real life.

	\begin{table}[!t]
		\renewcommand{\arraystretch}{1}
		\centering
  \caption{\small Conditional performances of kNN  and LocUNet.}
        \scalebox{0.7}{  
        \begin{tabular}{c|c|c|c|c|c|c}
			\hline	
			\rowcolor{Gray} \cellcolor{Grayyy} \bfseries No BS above NF:  &	1&	2&	3&	4&	5&\bfseries Overall\\
			\hline
			\bfseries kNN (k=200) & $69.81$ & $48.89$  & $34.62$ &$26.30$ & $19.62$& $27.38$\\
			\hline
			\bfseries kNN Cond. & $\mathbf{46.25}$ & $\mathbf{35.56}$  & $\mathbf{29.88}$ & $\mathbf{24.28}$ & $19.62$& $\mathbf{24.44}$\\
			\hline
			\bfseries kNN Cond. w/ below NF & $48.9$ & $42.1$  & $34.22$ & $26.45$ & $19.62$& $26.31$\\
			\hline
			\hline
			\bfseries LocUNet &$\mathbf{30.51}$	&$\mathbf{21.89}$& $\mathbf{16.31}$	& $\mathbf{13.28}$&	$9.62$& $\mathbf{13.14}$ \\
			\hline
			\bfseries LocUNet Cond.& $56$ &$28.37$& $20.1$	& $15.02$&	$\mathbf{9.06}$& $15.04$ \\
			\hline
			\bfseries LocUNet Cond. w/ below NF & $32.14$ &$39.03$& $18.45$	& $13.93$&	$\mathbf{9.06}$& $14.86$ \\
			\hline
			\bfseries LocUNet Cond. 5 & $33.57$ &$23.55$& $17.28$	& $13.5$&	$\mathbf{9.06}$& $13.3$ \\
			\hline
			\rowcolor{Grayy} \bfseries no instances in test set &$3284$	&$12478$ &$26515$	&$35167$ &$72556$ &$150000$\\
			\hline
			\rowcolor{Grayy}\bfseries  $\%$ &$2.19$	&$8.32$ &$17.68$	&$23.44$ &$48.37$ &$100$\\
			\hline
		\end{tabular}
		}
		 \label{table:condRunTime}
   \vspace{-8.2mm}
\end{table}

In Table \ref{table:condRunTime} we demonstrate the performance of LocUNet and kNN in the Robustness Scenario, conditioned on the number of BSs detectable at the UE.
We compare the performance of training (for kNN, setting the $k$ parameter) on the whole training set with that of training separately for different BS numbers. For the conditional training, we consider two variants, one which only uses the radio map estimation of a BS when the UE is in the coverage of the BS, and one which uses all the radio map estimations, even when a link between the UE and a BS is not strong enough for detection. We observe that such non-detectable signal information can be beneficial for LocUNet, while kNN demonstrates a better performance without taking such radio maps into consideration. We also observe that LocUNet performs the best when only one network with the whole training dataset is trained, whereas kNN achieves a better performance under conditional training. This can be attributed to the limited size of the dataset for low number of BSs (cf. bottom of the Table \ref{table:condRunTime}), causing LocUNet to overfit. In fact, the results with a network trained with only data from 5 BS in reception shows a deterioration with respect to training with the whole data. A detailed study of the effect of number of BSs in reception, dataset sizes, or neural network sizes could be interesting but is out of the scope of the current work.

    \vspace{-6.2mm}
	\subsection{Robustness to Out-of-Distribution Pathloss Measurements}\label{subseq:OOD}
 \begin{table}[!t]
		\renewcommand{\arraystretch}{1}
		\centering
  \caption{\small Neural networks of the Nominal and Robustness Scenarios evaluated under out-of-distribution pathloss measurements.  Comparison with kNN where $k$ parameter is fixed for the scenario of the interest. The accuracies under in-distribution measurements are shown in red.}
  \vspace{-2.2mm}
        \scalebox{0.8}{  
        \begin{tabular}{c|c|c|c|c}
        \hline	
		\rowcolor{Grayy} &\bfseries DPM &\bfseries DPM w/ cars &\bfseries IRT &\bfseries
		IRT w/ cars \\
		\hline
		\bfseries LocUNet Nominal & \textcolor{red}{4.80} & $\mathbf{8.80}$ & $20.07$ & $25.64$ \\
		\hline
		\bfseries kNN Nominal (k=16) & \textcolor{red}{6.95} & $11.29$ & $24.38$ & $28.56$ \\
		\hline
		\bfseries LocUNet Robustness & $12.94$ & $12.52$ & $\mathbf{11.93}$ & \textcolor{red}{13.14} \\
		\hline
		\bfseries kNN Robustness (k=200) & $9.22$ & $12.46$ & $23.41$ & \textcolor{red}{27.38} \\
        \hline
        \end{tabular}
		}
		 \label{table:OOD}
   \vspace{-8.2mm}
	\end{table}

 \vspace{-2.2mm}
	In this Section we show the performance of LocUNet and the kNN method for the above defined two scenarios under out-of-distribution (OOD) pathloss measurements, \textcolor{black}{i.e., we test with measurements stemming from different distributions than the one during the training.}
	
	To this end, we take the trained LocUNet networks under these two scenarios, and kNN method with $k$ values found via grid search in the training set for each scenarios. We then test both methods with pathloss measurements obtained from different simulations scenarios. In  addition to the above mentioned two ground-truth scenarios (DPM and IRT with cars), we consider two more scenarios. \textbf{1}) DPM with cars, where the radio map is simulated with the DPM under the presence of additional (unknown to LocUNet) obstructions (cars), \textbf{2}) IRT simulations as in the Robustness Scenario, but without the presence of unknown cars. The data for these scenarios are also available in the RadioLocSeer Dataset.
	In Table \ref{table:OOD} we observe that the network trained in the Nominal scenario suffers from substantial decrease of performance when the pathloss measurements stem from other scenarios, whereas the accuracy of the network trained in Robustness Scenario is not much affected, while being outperformed by the network trained in the Nominal Scenario when the measurements stem from DPM simulations, which demonstrates the importance of having high accuracy radio map estimations, when possible. We also notice that the network trained in Robustness Scenario performs even better when there are no unknown cars in the map. On the other hand, we observe that the performance of the kNN method is mainly dependent on the closeness of the real (surrogate) and the estimated maps, and different $k$ (nearest neighbors) values do not lead to remarkable improvements in performance, especially when the mismatch between the estimated and true radio map is high.

    \vspace{-6.2mm}
    \subsection{Robustness to Location Errors}\label{subseq:LocError}
    \vspace{-2.2mm}
    % In the following, we study the effect of different position uncertainties. 
    \subsubsection{Location Errors in the Training Dataset}\label{subseq:LocErrorPosTra}
    \begin{figure}[!t]
    \centering
    \includegraphics[width=0.45\textwidth]{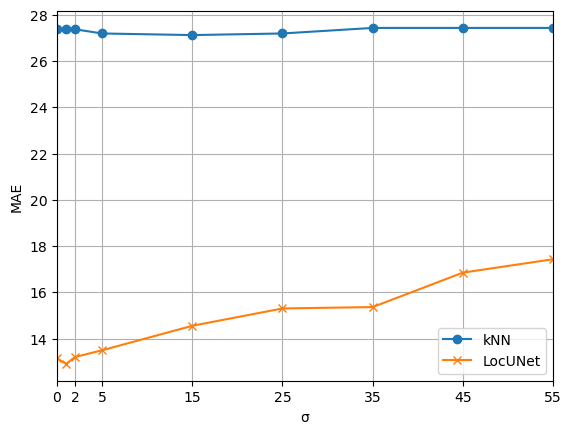}
    \vspace{-4.2mm}
    \caption{Robustness to the inaccuracies in the reported positions of the training dataset samples}
    \label{fig:NoisyLabel}
    \vspace{-4.2mm}
\end{figure}
 
    A crucial requirement of a localization method should be robustness to inaccuracies of the reported locations of the samples in the training dataset, as the very localization problem in dense urban setting, which the current paper is dealing with, is also present while collecting the labeled (here, the true locations) data.
    To assess the robustness of LocUNet to such inaccuracies, we consider additive Gaussian noise on each location coordinate of the training data. We report the test results for varying standard deviation of the noise in Fig. \ref{fig:NoisyLabel} and Table \ref{table:RobLocError}. We observe that even to a significantly high level of additive noise \textcolor{black}{($\sigma = 55$)} to the positions in the training data, both LocUNet and kNN \cite{RADAR} show extremely high robustness. \textcolor{black}{Another observation we made is a trend of decrease in the number of training epochs which yields the best validation loss with increasing $\sigma$, i.e., overfitting occurs earlier as $\sigma$ increases and the generalization is more severely impaired.} 

           \begin{table}[!t]
		\renewcommand{\arraystretch}{1}

		\centering
 \caption{\small \textcolor{black}{Test accuracies under training with contaminated position labels. The numbers of training epochs of LocUNet that give rise to the best validation losses are reported in the last row.}}
        \scalebox{0.8}{
        \begin{tabular}{c|c|c|c|c|c|c|c|c|c}
			\hline	
			\rowcolor{Grayy}  {\bfseries  Algorithm }& \bfseries $\sigma = 0$& \bfseries $\sigma = 1$& \bfseries $\sigma = 2$ &\bfseries $\sigma = 5$ & \bfseries $\sigma = 15$ & $\sigma = 25$ & $\sigma = 35$ & $\sigma = 45$ & $\sigma = 55$\\
			\hline
			 kNN \cite{RADAR}  & $27.38$ & $27.38$  & $27.38$   & $27.20$ & $27.13$ & $27.20$ & $27.44$ & $27.44$ & $27.44$\\
			\hline
   \hline
			\bfseries LocUNet &  $\mathbf{13.14}$ &  $\mathbf{12.90}$&$\mathbf{13.20}$  &$\mathbf{13.49}$ & $\mathbf{14.55}$ & $\mathbf{15.30}$ & $\mathbf{15.36}$ & $\mathbf{16.85}$& $\mathbf{17.43}$\\
			\hline
   \textcolor{black}{best $\#$ of epochs} &  $20$ &  ${17}$&${30}$  &${14}$ & $8$ & $8$ & $8$ & $9$& $6$\\
			\hline
		\end{tabular}
		}
		 \label{table:RobLocError}
   \vspace{-9.2mm}
	\end{table}
    \subsubsection{Uncertainty of Anchor BS Locations}
    Throughout the paper, we focused on a setting with known anchor BS locations, which are naturally known to the service providers, and might be also available to the users (required for self-localization at the UEs), through open access sources such as \url{https://www.cellmapper.net}. Here, we present a brief numerical study of LocUNet, when the locations of the anchor BSs are \emph{not} exactly known. As before, we model the uncertainty as an additive Gaussian random variable. Since generating a dataset for each realization of such uncertainty is unfeasible, we instead distort the position of the UE for each considered BS, and assume that the pathloss measurements stem from the distorted location (rounded so that it corresponds to a pixel in the radio map), while only the non-contaminated location is known. If the distorted location corresponds to a location in a building, we repeat the process of adding noise to the original location of the UE, until it does not. The results of LocUNet and kNN are reported in Table \ref{table:RobLocErrorTx}. In the first two rows, we show the accuracy of the algorithms trained with data with exactly known transmitter locations and evaluate their performances under contaminated transmitter locations (Zero-Shot Scenario, ZS). In the last two rows, the results under training with distorted BS locations are presented, where we observe the beneficial effects of it with respect to the Zero-Shot Scenario on the accuracy of the both methods.

% \begin{table}[!t]
% 		\renewcommand{\arraystretch}{1}

% 		\centering
%   \caption{\small Robustness to the inaccuracies in the reported positions of the training dataset samples. The results are in MAE.}
%         \scalebox{0.8}{
%         \begin{tabular}{c|c|c|c|c|c|c|c}
% 			\hline	
% 			\rowcolor{Grayy}  {\bfseries  Algorithm }& \bfseries $\sigma = 0$& \bfseries $\sigma = 2$ &\bfseries $\sigma = 5$ & \bfseries $\sigma = 10$ & $\sigma = 15$ & $\sigma = 20$ & $\sigma = 25$\\
% 			\hline
% 			LocUNet &  $\mathbf{13.14}$ &  $\mathbf{12.76}$&$\mathbf{13.47}$  &$\mathbf{13.90}$ & $\mathbf{13.79}$ & $\mathbf{14.18}$ & $\mathbf{13.58}$ \\
% 			\hline
% 			kNN \cite{RADAR} (k=200,200,300,400,400,500,700)  & $27.38$ & $27.38$   & $27.20$ & $27.13$ & $27.13$ & $27.12$ & $27.20$\\
% 			\hline
% 		\end{tabular}
% 		}
% 		 \label{table:RobLocError}
% \end{table}

    \begin{table}[!t]
		\renewcommand{\arraystretch}{1}

		\centering
  \caption{\small Robustness to the inaccuracies in the anchor BS locations. For each transmitter, the reported pathloss stems from a position contaminated by additive Gaussian noise with given standard deviation $\sigma$, to emulate the uncertainty in the BS position.}
        \scalebox{0.8}{
        \begin{tabular}{c|c|c|c|c|c}
			\hline	
			\rowcolor{Grayy}  {\bfseries  Algorithm } &\bfseries $\sigma = 5$ & \bfseries $\sigma = 10$ & $\sigma = 15$ & $\sigma = 20$ & $\sigma = 25$\\
			\hline
% 			kNN \cite{RADAR} Test Clean (k=500,800,6000,1000,12000)  &  $27.11$  & $27.26$ & $33.05$ & $36.60$ & $37.77$ \\
% 			\hline
% 			LocUNet Test Clean & $\mathbf{14.11}$  &$\mathbf{15.85}$ & $\mathbf{18.49}$ & $\mathbf{19.84}$ & $\mathbf{22.67}$ \\
% 			\hline	\hline
			kNN \cite{RADAR} Noisy Loc. ZS (k=200)  &  $30.91$  & $36.34$ & $43.67$ & $50.98$ & $57.95$ \\
			\hline
			LocUNet Noisy Loc. ZS& $\mathbf{16.42}$  &$\mathbf{22.23}$ & $\mathbf{29.01}$ & $\mathbf{35.29}$ & $\mathbf{40.96}$ \\
			\hline	\hline
			kNN \cite{RADAR} Noisy Loc. (k=500,800,6000,10000,12000)  &  $30.11$  & $34.78$ & $38.81$ & $42.54$ & $45.19$ \\
			\hline
			LocUNet Noisy Loc.& $\mathbf{16.26}$  &$\mathbf{20.52}$ & $\mathbf{25.43}$ & $\mathbf{29.55}$ & $\mathbf{34.09}$ \\
			
			\hline
		\end{tabular}
		}
		 \label{table:RobLocErrorTx}
   \vspace{-8.2mm}
	\end{table}

    \vspace{-6.2mm}
	
	\subsection{Comparison with ToA Ranging-Based Methods}\label{subseq:NumericalToA}
\vspace{-2.2mm}

% 	\begin{table}[!t]
% 	\renewcommand{\arraystretch}{1}
% 	\centering
%     \scalebox{0.8}{
% 	\begin{tabular}{c|c|c|c|c|c|c|c}
% 	\hline
% 		\rowcolor{Grayy}  {\bfseries  no BS:}&$\mathbf{5}$&$\mathbf{3}$&$\mathbf{5}$&$\mathbf{3}$&$\mathbf{5}$&$\mathbf{3}$ & {\bfseries Run-time}\\
% 		\hline	
% 		\rowcolor{Grayy}  {\bfseries  Standard dev. of noise:}&\multicolumn{2}{c|}{ $\sigma = 0$} & \multicolumn{2}{c|}{$\sigma = 10$} & \multicolumn{2}{c|}{$\sigma = 20$} & {\bfseries $\sim$ ms}\\
% 		\hline
% 		POCS \cite{POCSgholami2011wireless,POCSHero} & $37.75$ & $46.15$  & $39.39$ & $47.27$ & $41.34$ & $48.72$ & 15\\
% 		\hline		
% 		SDP  \cite{SDP}& $6.64$ & $16.63$ & $15.56$ & $30.06$  &  &  &  $600$\\
% 		\hline
% 		Robust SDP 1 \cite{SDPR}, $b=20$ &  &  &  &  &  &  &  $600$\\
% 		\hline
% 		Robust SDP 1 \cite{SDPR}, $b=0.7$ &  &  &  &  &  &  &  $600$\\
% 		\hline
% 		Robust SDP 2 \cite{ImpSDPR}, $b=20$ &  &  &  &  &  &  &  $600$\\
% 		\hline
% 		Robust SDP 2 \cite{ImpSDPR}, $b=0.7$ &  &  &  &  &  &  &  $600$\\
% 		\hline
% 		Bisection-based robust method \cite{BisecRob}, $b=20$ &  &  &  &  &  &  &  $16$\\
% 		\hline
% 		Bisection-based robust method \cite{BisecRob}, $b=0.7$ &  &  &  &  &  &  &  $16$\\
% 		\hline
% 		Max. correntropy criterion method \cite{Correntropy} &  &  &  &  &  &  &  $30$\\
% 		\hline\hline
% 		LocUNet Nominal $\&$ Robustness Scenarios &  &  &  &  &  &  &  5\\
% 	\end{tabular}
% 	}
% 	\caption{\small Comparison with ToA ranging-based methods subject to additive ToA measurement noise. } \label{table:ToAAllNF}
% \end{table}

    Using the RadioToASeer Dataset (Sec. \ref{subseq:Dataset_RadioToASeer}), we evaluate the accuracies of the state-of-the-art ToA ranging based algorithms, in our presented realistic urban setting, and compare with our method. The compared methods are based on the measured ToAs of the dominant paths of the beacon signals of BSs as explained in Section \ref{subseq:Dataset_RadioToASeer}, found in our RadioToASeer Dataset. 
    %The average lengths of the dominant paths in the RadioToASeer Dataset are 86.11m and 86.26m in the training and test sets, respectively. 
    We assumed the availability of the ToA measurements even if the signal power of the dominant ray of interest falls below the noise floor. This further favors ToA-based schemes. In a separate set of experiments not shown here for space limitation we have verified that if only the rays with RSS above the noise floor are considered (average of 4.72 BS above the noise floor), similar results are obtained with 2-3 meters of degradation in the performances of SDP-based methods, and around 1 meter of improvement in bisection-based methods.

    In practice, there are several factors which deteriorate the accuracy of ToA measurements at a UE, e.g., clock offset and skew (synchronization errors), quantization of the measured values, additional thermal noise, or failures in resolving the multipath components. In the numerical experiments, we only took into account the inaccuracy due to the quantization (in LTE, ToAs are quantized to intervals of 9.77 meters), the thermal noise and the clock error and modeled them altogether with a single additional Gaussian noise variable with zero mean and standard deviation $\sigma$, as proposed in \cite{probToA}. As in \cite{probToA}, we took $\sigma=20$ meters, and also evaluated the methods with $\sigma=10$m and under no measurement noise, $\sigma=0$ (cf. Table \ref{table:ToAAll}). For all the cases, all the 5 BSs or a subset of 3 BSs are taken into consideration.

\begin{table}[!t]
	\renewcommand{\arraystretch}{1}
	\centering
 \caption{\small Comparison with ToA ranging-based methods. Absolute Euclidean distance (MAE) accuracies and approximate run-times of compared algorithms under different additional noise and number of BSs in the map.}
    \scalebox{0.8}{
	\begin{tabular}{c|c|c|c|c|c|c|c}
	\hline
		\rowcolor{Gray}  {\cellcolor{Grayyy} \bfseries  no BS:}&$\mathbf{5}$&$\mathbf{3}$&$\mathbf{5}$&$\mathbf{3}$&$\mathbf{5}$&$\mathbf{3}$ & {\cellcolor{Grayy} \bfseries Run-time}\\
		\hline	
		\rowcolor{Gray}  {\cellcolor{Grayyy} \bfseries  Standard dev. of noise:}&\multicolumn{2}{c|}{ $\sigma = 0$} & \multicolumn{2}{c|}{$\sigma = 10$} & \multicolumn{2}{c|}{$\sigma = 20$} & {\cellcolor{Grayy}\bfseries $\sim$ ms}\\
		\hline
		POCS \cite{POCSgholami2011wireless,POCSHero} & $37.75$  & $46.15$ & $39.36$ & $47.26$ & $41.27$ & $48.72$ & 15\\
		\hline		
		SDP  \cite{SDP}& $\mathbf{6.81}$  & $\mathbf{13.95}$  & $15.52$  & $27.81$ & $24.88$ & $41.02$ & $600$\\
		\hline
		Robust SDP 1 \cite{SDPR}, $b=20$ & $9.98$ & $17.10$ & $17.02$ & $27.59$ & $27.76$ & $40.37$ & $600$\\
		\hline
		Robust SDP 1 \cite{SDPR}, $b=0.7$ & $7.04$ & $15.38$ & $17.13$ & $28.12$ & $28.42$ & $41.74$ & $600$\\
		\hline
		Robust SDP 2 \cite{ImpSDPR}, $b=20$ & $12.53$ & $19.63$ & $18.66$ & $28.59$ & $27.81$ & $39.93$ & $600$\\
		\hline
		Robust SDP 2 \cite{ImpSDPR}, $b=0.7$ & $7.14$ & $15.63$ & $17.10$ & $28.10$ & $28.39$ & $41.72$ & $600$\\
		\hline
		Bisection-based robust method \cite{BisecRob}, $b=20$ & $9.16$ & $15.87$ & $\mathbf{14.42}$ & $\mathbf{25.40}$ & $\mathbf{23.30}$ & $\mathbf{38.14}$ & $16$\\
		\hline
		Bisection-based robust method \cite{BisecRob}, $b=0.7$ & $9.49$ & $14.95$ & $14.80$ & $27.35$ & $24.09$ & $40.75$ & $16$\\
		\hline
		Max. correntropy criterion method \cite{Correntropy} & $12.44$ & $18.40$ & $20.54$ & $31.27$ & $31.91$ & $45.68$ & $30$\\
		\hline\hline
		LocUNet Nominal $\&$ Robustness Scenarios &  $\mathbf{4.80}$&$\mathbf{10.70}$ &  $\mathbf{13.14}$& $\mathbf{19.06}$ &  $\mathbf{13.14}$& $\mathbf{19.06}$ & 5\\
	\end{tabular}
	}
	 \label{table:ToAAll}
  \vspace{-9.2mm}
\end{table}

    For the robust ToA methods \cite{SDPR,ImpSDPR,BisecRob}, we pick the NLOS bias $b$ at CDF 0.9 (20 m), as suggested in \cite{Correntropy}, and also at CDF 0.5 (0.7 m) (cf. Fig. \ref{fig:eCDF}). 
    
	We report the simulation results in Table \ref{table:ToAAll}. We added the results of LocUNet in the Nominal and Robustness Scenarios, at the bottom of the Table, for comparison. The results of LocUNet under the Nominal and Robustness Scenarios should be compared with the ToA ranging-based methods under no additive measurement noise and  under noisy measurements, respectively. We observe that LocUNet outperforms all the compared ToA-based methods in all the settings.

\vspace{-6mm}
	\subsection{Comparison with Baselines in Image Processing}\label{subseq:Baselines}
\vspace{-2.2mm}			
	To demonstrate the validity of our method, we perform comparisons with baselines. We relate our method to the localization problems (e.g. human/hand pose estimation) in image processing, which can be generally characterized as a so-called \emph{keypoint} regression problem \cite{robustKeypoint}. 
	We describe in the following four approaches, with increasing level of accuracy, where our method falls into the last category. We report our findings based on the numerical experiments we conducted, and discuss the results.

%  \begin{figure}[!t]
% 		\centering

% % 		\begin{subfigure}{\textwidth}
% % 			\subfloat[][Nominal, 3 Tx]{\includegraphics[width=.33\textwidth]{hetmapRegVsNomTx3.png}} %\quad
% % % 			\subfloat[][Surrogate 1, 3 Tx]{\includegraphics[width=.33\textwidth]{hetmapRegVsSur1Tx3.png}} %\quad
% % 			\subfloat[][Surrogate 2, 3 Tx]{\includegraphics[width=.33\textwidth]{hetmapRegVsSur2Tx3.png}}\ %\quad%\\
% % 		\end{subfigure}
		
% 		\begin{subfigure}{\textwidth}
% 		\hspace*{\fill}%
% 			\subfloat[][Nominal Scenario]{\includegraphics[width=.43\textwidth]{hetmapRegVsNomTx5.png}}\hfill %\quad
% % 			\subfloat[][Surrogate 1, 5 Tx]{\includegraphics[width=.33\textwidth]{hetmapRegVsSur1Tx5.png}} %\quad
% 			\subfloat[][Robustness Scenario]{\includegraphics[width=.43\textwidth]{hetmapRegVsSur2Tx5.png}}\ %\quad%\\
% 			\hspace*{\fill}%
% 		\end{subfigure}
		
% 		\caption{\small Heatmap regression vs spatial regression (LocUNet) with varying standard deviation of the ground truth Gaussian heatmaps. Blue and orange curves show the performances of inference by CoM (mean) and by $\arg\max$ (mode), respectively. Constant green line shows the performance of LocUNet.} 
% 		\label{fig:heatmapVs}
%   \vspace{-5.2mm}
% 	\end{figure}

	\subsubsection{Direct classification}
\textcolor{black}{For the localization problem at hand, a classification approach seems to be the most appropriate at first sight. In this case, the task of the DNN is to approximate a one-hot map with the ``hot’’ pixel at the desired location. The classical loss function in this case is the cross-entropy loss between the desired one-hot map and the heatmap output by the DNN (both maps can be interpreted as probability distributions on the set of all possible locations). While this approach may be appropriate for classification of categorical data (e.g., handwriting of letters and numbers), it is inappropriate for localization since the cross-entropy loss does not capture the proximity of the estimated position in the Euclidean distance sense. In fact, our regression approach based on MAE or MSE is much more meaningful in this sense. Our implementation of the classification approach did not converge, as expected.}
	\subsubsection{Direct regression} 
		Another reasonable approach could be the direct regression of the pixel coordinates under a distance based loss, where instead of the CoM layer, a fully connected layer would be used as the last layer.  
		This flattening at the last layer impairs the spatial generalizability of the network. In particular, the translation equivariance, which is crucial for the localization problem, is lost.
		Our experiments with this approach did not converge either. Similar observations were made in some previous works, e.g. in \cite{EyeTracking}.

	\subsubsection{Heatmap regression}

  \begin{figure}[!t]
    \centering
    \includegraphics[width=0.85\textwidth]{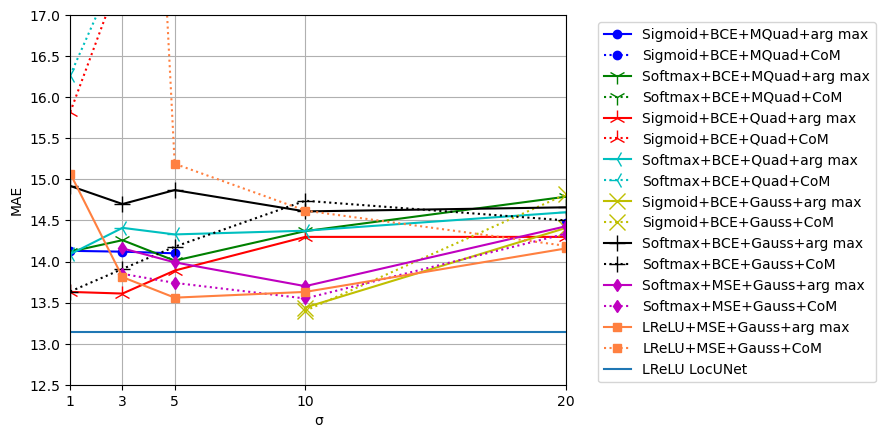}\vspace{-5.2mm}
    \caption{\textcolor{black}{Heatmap regression with different activations, heatmap loss functions, heatmap shapes vs LocUNet. Abbreviations:  \textbf{(Gauss)}: Gaussian, \textbf{(MQuad)}:  Inverse Multiquadric, \textbf{(Quad)}: Inverse Quadratic,  \textbf{(BCE)}: Pixelwise binary cross-entropy.}}
    \vspace{-6.2mm}
    \label{fig:heatMapResults}
\end{figure}	

\begin{table}[!t]	
	\renewcommand{\arraystretch}{1}
 
	\centering
 \caption{\textcolor{black}{5 best performing choices for heatmap regression vs LocUNet}}
 \vspace{-3.2mm}
	\scalebox{0.85}{
	\begin{tabular}{c|c|c}
		%\hline
		\rowcolor{Gray}		\bfseries Method &  \bfseries $\sigma$ & MAE\\
		%		\hline
	    Sigmoid+BCE+Gaussian+CoM & 10 & \bfseries 13.40\\
		%		\hline
		Sigmoid+BCE+Gaussian+argmax & 10 & 13.44\\
		%		\hline
		Softmax+MSE+Gaussian+CoM &  10 & 13.55\\
		%		\hline
		LeakyReLU+MSE+Gaussian+argmax & 5 & 13.56\\
		%		\hline
		Sigmoid+BCE+Inverse quadratic+argmax & 3 & 13.61\\
	\hline
        \hline
        LocUNet & - & \bfseries 13.14

	\end{tabular}
	}
	 \label{table:heatmapVs}
  \vspace{-8.2mm}
\end{table}

	To preserve the spatial information, the flattening of the layers has to be avoided.
	Heatmap regression based methods use fully convolutional networks to maintain the spatial information. 
 In such methods, the networks are trained to match \textcolor{black}{a (ground truth) heatmap, which are usually generated as 2D Gaussian distributions (standard deviation $\sigma$ being a hyperparameter to be tuned)} centered on the ground truth pixels. \textcolor{black}{The values of the heatmap for the  \emph{Gaussian} case is obtained by $e^{-\frac{r^2}{2\sigma^2}}$, where $r$ is the distance to the ground truth pixel from the pixel of interest. Additionally, we used two other radial basis functions for heatmap generation, namely, 
 \emph{Inverse Multiquadric}: $1/{\sqrt{1+{r^2}/{(2\sigma^2)}}}$ and \emph{Inverse Quadratic}: ${1}/{(1+{r^2}/{(2\sigma^2}))}$.}
 
After the neural network is trained, the position is inferred by taking either the pixel with the largest value in the heatmap (the \emph{mode}) or the center of mass (the {\em mean}), 
	as the heatmap can be interpreted as an a posteriori probability distribution of the UE position, and the mode and mean correspond to {\em maximum a posteriori} and {\em posterior mean} estimation, respectively.  Hence, these methods are trained against a loss which is different from the actual metric of interest - the Euclidean distance.
 Our experiments with this approach with the mentioned inference functions (i.e., $\arg\max$ and CoM), \textcolor{black}{and with different activation functions, heatmap loss functions, and heatmap shapes} have demonstrated worse yet close performances to LocUNet for specific values \textcolor{black}{of $\sigma$, which are shown in Fig. \ref{fig:heatMapResults} and summarized in Table \ref{table:heatmapVs}.}

 % In such methods, the networks are trained to match the ground truth heatmap, which are usually generated as 2D Gaussian distributions (variance being a hyperparameter to be tuned) centered on the ground truth pixels, and the position is inferred by taking either the pixel with the largest value in the heatmap (the \emph{mode}) or the center of mass (the {\em mean}), 
	% as the heatmap can be interpreted as an a posteriori probability distribution of the UE position, and the mode and mean correspond to {\em maximum a posteriori} and {\em posterior mean} estimation, respectively.  Hence, these methods are trained against a loss which is different from the actual metric of interest - the Euclidean distance. Our experiments with this approach with the mentioned inference functions (i.e., $\arg\max$ and CoM) have demonstrated worse yet close performances to LocUNet for specific variance values, shown in Fig. \ref{fig:heatmapVs}. 

	\subsubsection{Spatial regression} 
	This approach, which LocUNet is special case of, is based on relying on the heatmaps only as a latent representation, and defining the loss directly on the location coordinates. The inference of the coordinates are achieved by a differentiable function (mostly by CoM as in our paper), which allows end-to-end training. Equipped with the spatial generalization capability thanks to the convolutional network, together with the direct regression of the coordinates, such methods outperform the heatmap regression approach \cite{nibali2018numerical,robustKeypoint,laplaceLandmark,integral,luvizon}.
	
    \vspace{-5.2mm}
	
	\subsection{Ablation Study}\label{subseq:Ablation}
  \vspace{-2.2mm}
	
\begin{table}[!t]
		\renewcommand{\arraystretch}{1}
		\centering
  \caption{\small Performance of LocUNet in MAE and RMSE with different activation functions at the last layer before CoM, with different loss functions for training, and with different additional input features. n.c.: no convergence.}
  \vspace{-2.2mm}
        \scalebox{0.65}{  
        \begin{tabular}{c|c|c|c|c|c|c|c|c|c|c}
        \hline	
        \multicolumn{3}{c}{\cellcolor{Grayyy} \bfseries Training Loss:} & \multicolumn{4}{c|}{\cellcolor{Gray}\bfseries MAE} & \multicolumn{4}{c}{\cellcolor{Gray} \bfseries MSE}\\
        \hline
		\rowcolor{Grayy}\bfseries Inputs  &\bfseries $\#$ parameters &\bfseries Test Metric &\bfseries Leaky ReLU &\bfseries ReLU &\bfseries Softmax &\bfseries
		Sigmoid&\bfseries Leaky ReLU &\bfseries ReLU &\bfseries Softmax &\bfseries
		Sigmoid\\
		\hline
		\multirow{2}{*}{\bfseries wC wT} & \multirow{2}{*}{$22,332,381$} &\bfseries MAE &  $\mathbf{13.14}$ & n.c. & $13.25$ & $13.15$ & $\mathbf{14.45}$ & $15.41$ & $15.04$ & $15.04$\\
		\cline{3-11}
		 & &\bfseries RMSE & $\mathbf{21.78}$ & n.c.  & $23.53$ & $23.09$ & $\mathbf{21.93}$ & $22.81$ & $23.72$ & $22.35$\\
		\hline\hline
		\multirow{2}{*}{\bfseries w/oC wT} & \multirow{2}{*}{$22,331,676$} &\bfseries MAE&  $13.32$ & n.c.  & $\mathbf{12.88}$ & $13.47$ & $\mathbf{14.32}$ & n.c.  & $14.54$ & $14.66$\\
		\cline{3-11}
		 & &\bfseries RMSE & $23.66$ & n.c.  & $\mathbf{22.60}$ & $23.79$ & $21.98$ & n.c.  & $22.53$ & $\mathbf{21.88}$\\
		\hline\hline
		\multirow{2}{*}{\bfseries wC w/oT} & \multirow{2}{*}{$22,328,856$} &\bfseries MAE&   $\mathbf{12.86}$ & n.c.  & $13.00$ & $13.27$ & $14.75$ & n.c.  &  $\mathbf{14.45}$ & $14.91$\\
		\cline{3-11}
		 & &\bfseries RMSE & $\mathbf{22.26}$ & n.c.  & $23.27$ & $23.37$ & $\mathbf{21.95}$ & n.c.  & $22.07$ & $21.99$\\
		\hline 
		\hline
		\multirow{2}{*}{\bfseries w/oC w/oT} & \multirow{2}{*}{$22,328,151$} &\bfseries MAE &   $\mathbf{13.05}$ & n.c.  & $13.08$ & $13.36$ & $16.24$ & n.c.  & $\mathbf{14.40}$ & $14.69$\\
		\cline{3-11}
		 & &\bfseries RMSE & $\mathbf{22.37}$ & n.c.  & $22.55$ & $23.12$ & $23.44$ & n.c.  & $\mathbf{21.7}$ & $22.21$\\
		\hline
        \end{tabular}
		}
		 \label{table:activationInputsMSEMAE}
   \vspace{-7.2mm}
	\end{table}
	
		In this section, we discuss the effectiveness of our neural network design by comparing with other reasonable approaches. In the following, we present the impact of the additional input features, the choice of the last activation function of the network, and the choice of the loss function during the training. Resulting accuracies in MAE and root-mean-square error (RMSE) are shown in Table \ref{table:activationInputsMSEMAE}.

	\subsubsection{Impact of the Additional Input Features}\label{subsec:Impact_InpFeat}
		
	Our observation is that the impact of the additional input features (cf. Sec. \ref{subseq:Method}) on the performance is not significant. Hence, for the sake of simplicity of the exposition, throughout the paper we have taken all the additional features as inputs to the LocUNet, when comparisons with other methods are carried out. In Table \ref{table:activationInputsMSEMAE} we show the results of the numerical experiments with (\textbf{w}) and without (\textbf{w/o}) city (\textbf{C}) and Tx (\textbf{T}) maps as additional input features.
	
	\subsubsection{Impact of the Activation Function Before the CoM Layer}
	
	All of the above mentioned previous works in localization forced the heatmaps by the activation functions to become non-negative, or normalized them to represent a probability distribution (e.g. by Softmax as the activation function). Interestingly, our experiments with several activation functions (ReLU, Softmax, Sigmoid and Leaky ReLU), have shown, that allowing negative values in the heatmap (by using Leaky ReLU as the activation function), performed generally the best among all the activation functions (cf. Table \ref{table:activationInputsMSEMAE}). Here, this quasi-heatmap can be interpreted as a quasi-probability and the CoM as then the mean under a quasi-probability distribution, which is the MMSE estimator (See also the Examples in Sec. \ref{sec:Examples} and the \emph{belief/disbelief map} interpretation therein). A study of this interesting observation is, however, beyond the scope of this paper.

	\subsubsection{Impact of the Loss Function in the Training}

	Based on the results in the Table \ref{table:activationInputsMSEMAE}, we compare the influence of the training loss function on the evaluation metrics of MAE and RMSE in Table \ref{table:ImprovPerc}, where we show the improvement (in percentage) in MAE (RMSE) when trained with MAE (MSE) loss, with respect to when trained with MSE (MAE) loss. In mathematical terms, we calculate $\textup{Imp}_\textup{RMSE} = 100 \times \frac{\textup{RMSE}_\textup{MAE}-\textup{RMSE}_\textup{MSE}}{\textup{RMSE}_\textup{MAE}}$, $\textup{Imp}_\textup{MAE} = 100 \times \frac{\textup{MAE}_\textup{MSE}-\textup{MAE}_\textup{MAE}}{\textup{MAE}_\textup{MSE}}$, where the terms in fractions denote the performance in terms of $\textup{TestMetric}_\textup{TrainingLoss}$.

	We observe that in most of the cases training with MAE loss yielded more gains in MAE in comparison to the performance gains achieved in RMSE when trained with MSE. 

 \begin{table}[!t]
		\renewcommand{\arraystretch}{1}
		\centering
  \caption{\small Improvements (in percentage) observed in Table \ref{table:activationInputsMSEMAE} in MAE, RMSE metrics by training with MAE and MSE losses, with respect to training with MSE and MAE losses, respectively.}
        \scalebox{0.8}{  
        \begin{tabular}{c|c|c|c|c}
        \hline	
		\rowcolor{Grayy}\bfseries Inputs &\bfseries Improvement &\bfseries Leaky ReLU  &\bfseries Softmax &\bfseries Sigmoid\\
		\hline
		\multirow{2}{*}{\bfseries wC wT} & MAE  & $\mathbf{ 9.07\%}$ & $\mathbf{11.9}\%$ &  $\mathbf{12.58\%}$\\
		\cline{2-5}
		& RMSE & $-0.69\%$  & $-0.81\%$ & $3.21\%$\\
		\hline
		\hline
		\multirow{2}{*}{\bfseries w/oC wT} & MAE & $6.98\%$ & $\mathbf{11.42\%}$ & $\mathbf{8.10\%}$\\
		\cline{2-5}
		& RMSE & $\mathbf{7.10\%}$ & $0.31\%$ &  $8.03\%$\\
		\hline
		\hline
		\multirow{2}{*}{\bfseries wC w/oT} & MAE & $\mathbf{12.80\%}$ & $\mathbf{10.03\%}$ & $\mathbf{11.00\%}$\\
		\cline{2-5} 
		& RMSE & $1.39\%$ & $5.15\%$ & $5.91\%$\\
		\hline 
		\hline
		\multirow{2}{*}{\bfseries w/oC w/oT} & MAE  &  $\mathbf{19.65\%}$ & $\mathbf{9.17\%}$ &  $\mathbf{9.03\%}$\\
		\cline{2-5}
		& RMSE  & $-4.78\%$ & $3.7\%$ &  $3.94\%$\\
		\hline
        \end{tabular}
		}
		 \label{table:ImprovPerc}
   \vspace{-8.2mm}
\end{table}
    \vspace{-5.2mm}
	\section{Conclusions}\label{seq:Conclusions}
\vspace{-2.2mm}	
	
	In this paper we presented a deep learning method to localize a UE based on the measured signal strengths from a set of BSs. Our method, based on the estimates of the true radio maps of each BS, is tailored to work in realistic propagation environments in urban settings. The proposed approach does not necessitate currently uncommon hardware neither at UEs nor at BSs, and does not require an extensive measurement campaign, and thus can be easily integrated in the localization practices that are already in use. Our method outperforms the state-of-the-art localization algorithms in realistic urban scenarios and enjoys high robustness to inaccuracies in the radio map estimates, and also to inaccuracies in the positions labels of the training set. We also presented two new datasets, which can be used to evaluate the performances of pathloss fingerprint and ToA ranging-based algorithms in realistic urban environment scenarios.

	\vspace{-5.2mm}
	\section{Acknowledgment}
\vspace{-2.2mm}	
	We thank the anonymous referee for the careful and critical review of our work. The work presented in this paper was partially funded by the DFG Grant DFG SPP 1798 “Compressed Sensing in Information Processing” through Project Massive MIMO-II, and by the German Ministry for Education and Research as BIFOLD (ref. 01IS18037A). The authors are grateful to the HPC-Cluster of the Institute of Mathematics of the TU Berlin for the computing resources.
	
	%\newpage
 \vspace{-0.2mm}
	\bibliographystyle{IEEEtran}
	%\clearpage
	%\setstretch{1}
% 	{\footnotesize \bibliography{pub}}
    \vspace{-6.2mm}
 	\bibliography{pub}

\end{document}

%% file: header.tex
\usepackage{amsfonts, amscd, mathrsfs, amsopn, bbm, bbold, multicol, relsize}
\usepackage{mathrsfs}
\usepackage[psamsfonts]{amssymb}
\usepackage{mathbbol}
\usepackage[mathscr]{eucal}
\usepackage[cmex10]{amsmath}
\usepackage[amsmath,thmmarks]{ntheorem}
\usepackage{mathtools}
\usepackage{cite}
\usepackage{tikz}
\usepackage{graphicx}
\theoremstyle{plain}
\theoremheaderfont{\itshape\bfseries}
\theorembodyfont{\normalfont\itshape}
\theoremseparator{:}
\theoremsymbol{}%\vspace{0.2cm}}

\newtheorem{remark}{Remark}

\theoremstyle{nonumberplain}

\theoremstyle{nonumberplain}
\theoremheaderfont{\itshape}
\theorembodyfont{\normalfont}
\theoremseparator{:}
\theoremsymbol{}

\theoremstyle{plain}
\theoremheaderfont{\itshape\bfseries}
\theorembodyfont{\normalfont}
\theoremseparator{:}

\theoremstyle{nonumberplain}
\theoremsymbol{\rule{1ex}{1ex}}

\newlength\fheight
\newlength\fwidth
\def\eps{ \epsilon }

\newcommand\restr[2]{{% we make the whole thing an ordinary symbol
  \left.\kern-\nulldelimiterspace % automatically resize the bar with \right
  #1 % the function
  \right|_{#2} % this is the delimiter
  }}